\documentclass[5p,times]{elsarticle}

\usepackage[english]{babel}
\usepackage{times} 

\usepackage[utf8]{inputenc}
\usepackage{hyperref}
\usepackage[nolist]{acronym}

\usepackage{url}
\usepackage{textcomp}
\usepackage{footmisc}
\usepackage{booktabs}
\usepackage{flushend}
\usepackage{wrapfig} 

\usepackage{siunitx}
\usepackage{graphicx}
\usepackage{epsfig} 
\usepackage{pgfplots}
\usepackage{caption}
\usepackage[font=footnotesize]{subcaption}
\usepackage{pgfplots}

\usepackage[export]{adjustbox}

\usepackage{tikz}
\pgfdeclarelayer{foreground}
\pgfsetlayers{background,main,foreground}
\usetikzlibrary{quotes,angles,backgrounds,arrows,automata,shapes,positioning,calc,through,spy}

\tikzset{
    imglabel/.style={
      rectangle,
      inner sep=2pt,
      text=black,
      minimum height=1em,
      text centered,
      fill=white,
      fill opacity=1.0,
      text opacity=1,
      anchor=south west,
    },
  }

\tikzset{
	state/.style={
		rectangle,
		draw=black, very thick,
		minimum height=1.0em,
		text centered,
	},
}

\usepackage{algorithm}
\usepackage[noend]{algpseudocode}

\makeatletter
\def\BState{\State\hskip-\ALG@thistlm}
\makeatother

\usepackage{xcolor}
\definecolor{myYellow}{rgb}{0.93,0.69,0.13}
\definecolor{myPurple}{rgb}{0.49,0.18,0.56}
\definecolor{myGreen}{rgb}{0.26 0.72 0.54}

\newcommand{\blue}[1]{
	{\color{blue}{#1}}%
}

\newcommand{\red}[1]{
	{\color{red}{#1}}%
}

\usepackage{mathtools}
\usepackage{amsmath}
\usepackage{amssymb}
\usepackage{amsfonts}
\usepackage{nicefrac}
\usepackage{bm} 

\DeclareMathOperator*{\minimize}{minimize}
\DeclareMathOperator*{\maximize}{maximize}

\DeclareSIUnit{\litre}{l}

\usepackage{etoolbox}
\makeatletter%
\AfterPreamble{%
	\usepackage{hyperref}%
	\let\oldhypertarget\hypertarget%
	\renewcommand{\hypertarget}[2]{%
		\oldhypertarget{#1}{#2}%
		\protected@write\@mainaux{}{%
			\string\expandafter\string\gdef%
			\string\csname\string\detokenize{#1}\string\endcsname{#2}%
		}%
	}%
	\newcommand{\myhyperlink}[1]{%
		\hyperlink{#1}{\csname #1\endcsname}%
	}%
}
\makeatother%

\newcounter{Remark}
\newcommand{\displayRemarks}[2][]{%
	\stepcounter{Remark}%
	\textit{Remark~}\hypertarget{#1}{\theRemark}\textit{~(#2)}%
}

\newcommand{\refRemarks}[1][]{%
	Remark~\myhyperlink{#1}%
}

\newcounter{Definition}
\newcommand{\displayDefinitions}[2][]{%
	\stepcounter{Definition}%
	\textit{Definition~}\hypertarget{#1}{\theDefinition}\textit{~(#2)}%
}

\newcommand{\refDefinitions}[1][]{%
	Def.~\myhyperlink{#1}%
}

\newcounter{Problem}
\newcommand{\displayProblem}[2][]{%
	\stepcounter{Problem}%
	\textit{Problem~}\hypertarget{#1}{\theProblem}\textit{~(#2)}%
}

\newcommand{\refProblems}[1][]{%
	Problem~\myhyperlink{#1}%
}

\usepackage{scalerel}
\usetikzlibrary{svg.path}
\definecolor{orcidlogocol}{HTML}{A6CE39}
\tikzset{
	orcidlogo/.pic={
		\fill[orcidlogocol] 
		svg{M256,128c0,70.7-57.3,128-128,128C57.3,256,0,198.7,0,128C0,57.3,57.3,0,128,0C198.7,0,256,57.3,256,128z};
		\fill[white] svg{M86.3,186.2H70.9V79.1h15.4v48.4V186.2z}
		svg{M108.9,79.1h41.6c39.6,0,57,28.3,57,53.6c0,27.5-21.5,53.6-56.8,53.6h-41.8V79.1z 
		M124.3,172.4h24.5c34.9,0,42.9-26.5,42.9-39.7c0-21.5-13.7-39.7-43.7-39.7h-23.7V172.4z}
		svg{M88.7,56.8c0,5.5-4.5,10.1-10.1,10.1c-5.6,0-10.1-4.6-10.1-10.1c0-5.6,4.5-10.1,10.1-10.1C84.2,46.7,88.7,51.3,88.7,56.8z};
	}
}

\newcommand\orcidicon[1]{\href{https://orcid.org/#1}{\mbox{\scalerel*{
    \begin{tikzpicture}[yscale=-1,transform shape]
        \pic{orcidlogo};
    \end{tikzpicture}
}{|}}}}

\newcommand\copyrighttext{%
    \small \begin{center} \color{red} \textcopyright\,2025 Elsevier. Accepted for Robotics and Autonomous Systems. Personal use of this material is permitted. Permission from Elsevier must be obtained for all other uses, in any current or future media, including reprinting/republishing this material for advertising or promotional purposes, creating new collective works, for resale or redistribution to servers or lists, or reuse of any copyrighted component of this work in other works. \end{center}}
\newcommand\copyrightnotice{%
	\begin{tikzpicture}[remember picture,overlay]
	\node[anchor=south,yshift=26.8cm] at (current page.south) 
	{\color{red}\fbox{\parbox{\dimexpr\textwidth-\fboxsep-\fboxrule\relax}{\copyrighttext}}};
	\end{tikzpicture}%
}

\journal{Robotics and Autonomous Systems}

\begin{document}\sloppy

\begin{frontmatter}



\title{A Signal Temporal Logic Approach for Task-Based Coordination of Multi-Aerial Systems: a Wind Turbine Inspection Case Study}



\author[ctu,rse,corr]{Giuseppe Silano}

\author[use]{Alvaro Caballero}

\author[unisannio,enea]{Davide Liuzza}

\author[unisannio]{Luigi Iannelli}

\author[uza]{Stjepan Bogdan}

\author[ctu]{Martin Saska}

\address[ctu]{Department of Cybernetics, Faculty of Electrical Engineering, Czech Technical University in Prague, Prague, Czech Republic (emails: {\tt \{giuseppe.silano, martin.saska\}@fel.cvut.cz}).}

\address[use]{Department of Power Generation Technologies and Materials, Ricerca sul Sistema Energetico (RSE) S.p.A., Milan, Italy.}

\address[use]{GRVC Robotics Laboratory, University of Seville, Seville, Spain (email: {\tt alvarocaballero@us.es}).}

\address[unisannio]{Department of Engineering, University of Sannio in Benevento, Benevento, Italy (emails: {\tt \{davide.liuzza, luigi.iannelli\}@unisannio.it}).}

\address[enea]{Fusion and Technology for Nuclear Safety and Security Department, Italian National Agency for New Technologies, Energy and Sustainable Economic Development (ENEA), Frascati, Italy.} 

\address[uza]{Faculty of Electrical Engineering and Computing, University of Zagreb, Zagreb, Croatia (email: {\tt stjepan.bogdan@fer.hr}).} 
    
\address[corr]{Corresponding author}


\begin{acronym}
    \acro{CBF}[CBF]{Control Barrier Function}
    \acro{CVRP}[CVRP]{Capacity Vehicle Routing Problem}
    \acro{ILP}[ILP]{Integer Linear Programming}
    \acro{LP}[LP]{Linear Programming}
    \acro{LSE}[LSE]{Log-Sum-Exponential}
    \acro{LTL}[LTL]{Linear Temporal Logic}
    \acro{MILP}[MILP]{Mixed-Integer Linear Programming}
    \acro{MRS}[MRS]{Multi-Robot System}
    \acro{MTL}[MTL]{Metric Temporal Logic}
    \acro{NLP}[NLP]{Nonlinear Programming}
    \acro{PSO}[PSO]{Particle Swarm Optimization}
    \acro{ROS}[ROS]{Robot Operating System}
    \acro{SIL}[SIL]{Software-in-the-loop}
    \acro{STL}[STL]{Signal Temporal Logic}
    \acro{UAV}[UAV]{Unmanned Aerial Vehicle}
    \acro{TL}[TL]{Temporal Logic}
    \acro{TS}[TS]{Transition System}
    \acro{TSP}[TSP]{Traveling Salesman Problem}
    \acro{TWTL}[TWTL]{Time Window Temporal Logic}
    \acro{VRP}[VRP]{Vehicle Routing Problem}
    \acro{wSTL}[wSTL]{weighted-STL}
    \acro{wrt}[w.r.t.]{with respect to}
\end{acronym}



\begin{abstract}

The paper addresses task assignment and trajectory generation for collaborative inspection missions using a fleet of multi-rotors, focusing on the wind turbine inspection scenario. The proposed solution enables safe and feasible trajectories while accommodating heterogeneous time-bound constraints and vehicle physical limits. An optimization problem is formulated to meet mission objectives and temporal requirements encoded as \acf{STL} specifications. Additionally, an event-triggered replanner is introduced to address unforeseen events and compensate for lost time. Furthermore, a generalized robustness scoring method is employed to reflect user preferences and mitigate task conflicts. The effectiveness of the proposed approach is demonstrated through MATLAB and Gazebo simulations, as well as field multi-robot experiments in a mock-up scenario.

\end{abstract}










\end{frontmatter}



\copyrightnotice

\section{Introduction}
\label{sec:introduction}

The increasing use of~\acp{UAV} for civilian infrastructure inspections has paved the way for aerial vehicles with greater autonomy, thereby enhancing the safety, speed, and accuracy of these inspections~\cite{Ollero2024BookChapter}. The capacity for repeated operations enables the monitoring of infrastructure changes over time, providing benefits to a variety of applications such as inspecting oil and gas pipelines~\cite{OlleroRAM2018}, wind turbine blades~\cite{Car2020IEEEAccess}, power transmission lines~\cite{CaballeroIEEEAccess2023}, towers and bridges~\cite{Shakhatreh2019Access}.

Presently, the most widespread approach is the use of~\acp{UAV} that are controlled by highly trained operators. These experts are knowledgeable about the required information and site-specific challenges, such as obstacles, weather conditions, and time of day. On-board cameras and sensors are utilized to gather data which is then processed by specialist teams for further analysis. However, this approach has two major limitations: First, flight operations can be hazardous for operators who must maintain continuous visual contact with the~\acp{UAV}, especially in environments at height; Second, the process is time-consuming, costly, and prone to human error~\cite{Madridan2021ESA, Merkert2020JATM, SilanoICUAS2023}. Over time, various methods for automating the inspection task have been proposed in the literature~\cite{Shakhatreh2019Access, Car2020IEEEAccess, CaballeroIEEEAccess2023, OlleroRAM2018, Ollero2024BookChapter, KanellakisIEEEAccess2020}. Despite this, significant challenges still exist in making inspections autonomous, including the reliability of navigation systems, radio interference, limited battery life, and unpredictable environmental events, which pose risks to mission success~\cite{Langaker2021IJARS, BonillaIEEEComm}.

Therefore, there is a clear need for safe and effective techniques that can enhance maintenance and inspection operations, reduce risks, and decrease costs for companies. Formal methods~\cite{Pola2019ARC} can address these challenges by offering a concise way to express complex mission specifications and time constraints by leveraging their similarity to natural language commands and the use of connectivity operators. Specifically, constraints and objectives are encapsulated within a single temporal formula. 
Notably,~\acf{STL}~\cite{Maler2004FTMA}, particularly in its robust formulation~\cite{Donze2010FMATS}, employs the concept of quantitative semantics, meaning that it not only verifies that the system execution satisfies desired requirements, but also provides a metric of how well these requirements are met, referred to as \textit{robustness}~\cite{Donze2010FMATS}. This results in an optimization problem aimed at achieving high robustness scores values through the generation of a trajectory that satisfies the desired specifications while ensuring feasibility within dynamic and constraint margins.

\begin{figure}[tb]
    \centering
    \adjincludegraphics[width=0.95\columnwidth]{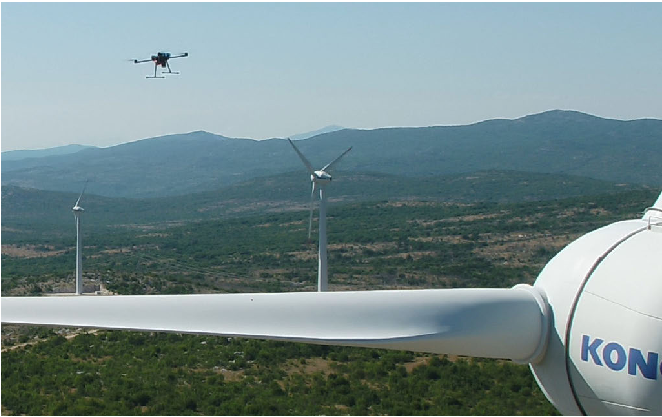}
    \caption{A multi-rotor \ac{UAV} conducting an inspection of a wind turbine, capturing videos and pictures of the nacelle, rotor shaft and blades, as well as their surrounding environment, to perform a preliminary remote evaluation \cite{Car2020IEEEAccess}.}
    \label{fig:windTurbineInspectionExample}
\end{figure}

In this paper, wind turbine inspection (see Figure~\ref{fig:windTurbineInspectionExample}) serves as a case scenario for presenting a multi-\ac{UAV} motion planning approach leveraging on~\ac{STL} specifications for collaborative inspection missions.
This scenario illustrates sophisticated planning, involving the coordination of multi-rotor~\acp{UAV} to avoid collisions with each other and the environment, accomplish diverse time-bound inspection tasks, adhere to the vehicle's dynamical constraints, and ensure comprehensive infrastructure coverage. The mission requirements are formulated as an~\ac{STL} formula, and the robustness of the formula is optimized by setting up a nonlinear, non-convex max-min optimization problem. To manage the complexity of this nonlinear optimization, a hierarchical approach is employed, starting with solving a~\ac{MILP} planning problem and subsequently feeding the final~\ac{STL} optimizer.

\subsection{Related work}
\label{sec:realtedWork}

Extensive research has been done in the fields of autonomous navigation and coordination capabilities for~\acp{UAV}~\cite{Shakhatreh2019Access, Madridan2021ESA, Merkert2020JATM, Langaker2021IJARS}. The primary objectives of this research have been to: (i) enable~\acp{UAV} to navigate autonomously and reach safe conditions while avoiding unsafe behaviors (e.g., collisions, crashes), and (ii) enhancing the vehicles' ability to coordinate for data collection and executing appropriate actions. Accurate trajectory planning and task assignment are crucial in both of these areas, not only for individual vehicles, but also for fleets of~\acp{UAV} with varying characteristics and capabilities that must cooperate, avoid obstacles, and meet mission requirements simultaneously in the same operational area.

The literature on~\textit{autonomous navigation} presents a diversity of point-to-point navigation approaches~\cite{Mansouri2018CEP}, including potential field methods~\cite{Tan2021Access} and vehicle-routing problem formulations~\cite{Nekovar2021RAL}. Many studies prioritize drone mission endurance to maximize coverage area by accounting for flight times. However, these approaches frequently overlook drone dynamics and physical constraints, resulting in infeasible paths. The problem then becomes a combinatorial optimization challenge, as exemplified by the well-known \ac{VRP}~\cite{Nekovar2021RAL}, which  becomes increasingly unsolvable within a reasonable time frame as the complexity exponentially increases with the number of vehicles and variables. In such cases, heuristic methods are often employed to simplify the complexity and identify the most viable solution.

Research has explored various approaches for endowing~\acp{UAV} with \textit{coordination capabilities}~\cite{Park2020ICRA, Tehrani2022IROS, Zhu2019RAL, Honig2018TRO}. Some studies have focused on centralized multi-agent algorithms for coordinating and planning safe and dynamically feasible trajectories~\cite{Park2020ICRA}. Others propose methods to convert the multi-agent task allocation problem into a search space that can be solved using the \ac{PSO} algorithm~\cite{Tehrani2022IROS}. Additionally, some approaches employ knowledge transfer techniques between agents to achieve collision avoidance and safety~\cite{Zhu2019RAL}. These approaches can efficiently coordinate and compute trajectories for multiple agents~\cite{Zhu2019RAL, Honig2018TRO}, but do not guarantee that vehicles will complete tasks within specified time windows. Furthermore, encoding possibly different complex temporal requirements into the optimization problem, such as a requirement for multiple drones to visit designated regions at specific times with certain behaviors while ensuring safety, requires a systematic approach. 

In addressing the motion planning challenge for multi-robot systems, various approaches have been explored in existing literature~\cite{Honig2018TRO, PantCDC2017, LuisRAL2020}. Many solutions rely on abstract agent dynamics~\cite{Honig2018TRO} or abstract grid-based environment representations~\cite{PantCDC2017}, integrating a discrete planner with a continuous motion generator. While these methods can swiftly compute collision-free motions for numerous agents, they fail to guarantee adherence to physical constraints or task completion within specified time frames. Additionally, they do not offer velocity and acceleration references, leaving the controller responsible for generating these signals. Conversely, when considering multi-rotor models~\cite{LuisRAL2020}, solutions depend on information exchange among agents, posing difficulties in environments with electromagnetic interference, such as in wind turbine inspection tasks~\cite{Krich2017TAES}. 

Recently, several studies have explored planning and coordination techniques that incorporate advanced specifications and temporal goals~\cite{Lindemann2020TCNS, Chen2012TRO, Leahy2021TRO}. However, these problems often prove challenging due to their nonlinear, non-convex optimization max-min nature.~\acp{CBF}~\cite{Lindemann2020TCNS} offer potential for finding robust and computationally efficient trajectories, though they are limited to simpler scenarios and lack soundness and completeness for the~\ac{STL} syntax. For example, if two robots need to visit distinct regions within overlapping time windows,~\acp{CBF} may lead to infeasible problems~\cite{Buyukkocak2022ECC}. Automata-based methods~\cite{Chen2012TRO} can reduce complexity by assigning specifications to individual agents. However, constructing the required transition system is a significant burden. Sampling-based techniques~\cite{Leahy2021TRO}, encode~\ac{STL} specifications as linear and boolean constraints, offering a solution to the challenge of finding optimal solutions. However, these techniques typically assume that each robot is specialized in specific skills.  
Most of these methods end up with a dynamic programming formulation, which suffer from dependency on the initial solution~\cite{Bertsekas2009Book}.



\subsection{Contributions}
\label{sec:contributions}

In this paper, we introduce a novel approach to multi-\ac{UAV} task assignment and trajectory generation for collaborative inspection missions, with wind turbine inspection serving as our case study. Our method utilizes~\ac{STL} as a specification language to express the mission's objectives and time constraints. These objectives include collision avoidance between the~\acp{UAV} and the environment, as well as the completion of time-bound inspection tasks, such as reaching specified target areas and ensuring comprehensive coverage of the infrastructure. Addressing these requirements results in a complex nonlinear, non-convex max-min optimization problem typically solved using dynamic programming. However, finding an optimal solution in a reasonable time frame can be challenging due to the solvers' tendency to get stuck in local optima based on the initial guess~\cite{Bertsekas2009Book}.
To tackle this challenge, we propose a two-step hierarchical approach that initially simplifies the mission into a~\acl{MILP} planning problem formulated on a subset of the inspection objectives and subsequently seeds the global~\ac{STL} optimizer. The approach extends our prior work~\cite{SilanoRAL2021} by:
\begin{itemize}
    \item Addressing collaborative inspection mission complexities and computing dynamically feasible trajectories with diverse time bounds and vehicle constraints (Section~\ref{sec:specificationMapping}). 
    \item Introducing a new method for computing the initial guess solution (Section~\ref{sec:milpEncoding}) considering heterogeneous constraints on vehicle velocity and acceleration.
    \item Incorporating an event-triggered replanner (Section~\ref{sec:eventBasedReplanner}) to modify the planned trajectory in case of disturbances or unexpected events, ensuring the new trajectory aligns with the previous optimal solution by compensating for lost time.
    \item Including a weighted generalization of the~\ac{STL} robust semantics~\cite{MehdipourLCSS2021} to capture user preferences and address issues arising from conflicting tasks when mission accomplishment cannot be achieved (Section~\ref{sec:attrititionAwarePlanner}).
    \item Assessing the method's overall performance through MATLAB simulations to evaluate its effectiveness in meeting mission specifications and the efficiency of incorporating the proposed initialization procedure (Section~\ref{sec:experimentalResults}). Additionally, validation is conducted via Gazebo simulations \cite{Dimmig2024RAM} and field experiments in a mock-up setting (Section~\ref{sec:fieldExperiments}).
\end{itemize}

The advantages our approach are twofold. Firstly, we employ an~\ac{MILP} planning formulation that operates on a simplified version of the problem, eliminating the need for linear or linearizable constraints and system dynamics, thereby facilitating the search for a global solution. Secondly, we utilize a concise and unambiguous~\ac{STL} formulation, enabling end-users to specify drone behavior in easily understandable terms, such as target areas to inspect and infrastructure zones to cover, while considering for explicit time requirements. In contrast to existing solutions~\cite{Lindemann2020TCNS, Chen2012TRO, Leahy2021TRO}, our approach utilizes a smooth robustness function to incorporate a quantitative robust semantics, a feature not achievable with automata-based methods~\cite{Chen2012TRO}. Furthermore, our approach addresses the issue of overestimating the dimensions of the robot and obstacles, which can lead to more conservative maneuvers with~\acp{CBF}~\cite{Lindemann2020TCNS}.

Also, one significant advantage of our approach using \ac{STL} over point-to-point navigation and vehicle-routing problem formulations, as listed in Section~\ref{sec:realtedWork}, is its ability to encapsulate intricate temporal constraints and requirements in a formal and expressive manner. By integrating natural language commands, temporal and Boolean operators, and task and motion planning, we can devise trajectories that fulfill mission objectives while considering the dynamics and physical constraints of the multi-rotor systems. Moreover, the proposed method allows for handling coordination among multiple drones to ensure the completion of all required inspections without specifying the task (the paper presents a general method applied to a particular case scenario) that a single vehicle must accomplish. Existing methods for planning from \ac{STL} specifications include abstraction-based methods, mixed-integer optimization, and nonlinear optimization. Abstraction-based methods \cite{Plaku2016AI} construct a discrete abstraction of the continuous state space, but suffer from scalability issues. Mixed-integer optimization methods \cite{Sun2022RAL, Yang2020ACC}  provide soundness and completeness (see Section~\ref{sec:smoothApproximation}) guarantees but become intractable with longer planning horizons. Nonlinear optimization approaches \cite{PantACM2018, Vasile2017IROS} offer increased generality and scalability but are prone to local optima based on initial guesses. Our two-step hierarchical approach addresses these challenges by initially simplifying the mission into a \ac{MILP} planning problem and then seeding the global \ac{STL} optimizer.



\section{Problem Description}
\label{sec:problemDescription}

This paper explores the problem of utilizing a fleet of multi-rotor~\acp{UAV} to collaboratively perform inspection missions, with a focus on wind turbine inspection as a case study. This entails capturing videos and pictures of the wind turbine and its surroundings to conduct a preliminary remote evaluation. 

The inspection procedure is divided into two distinct tasks: pylon inspection and blade inspection. The \textit{pylon inspection} involves visiting designated areas of interest, modeled as 3D spaces, to assess the structural integrity of mechanical components, such as the nacelle and rotor shaft. The objective of this inspection is to identify potential hazards, like corrosion or damage, that may affect the stability and safety of the pylon. In contrast, the \textit{blade inspection} requires obtaining scans that cover the entire surface of the blade. This need arises due to external factors, such as wind, rain, hail, or bird strikes, that can cause cracks or coating damage, potentially impacting the blades' performance. To ensure inspection accuracy, the~\acp{UAV} must maintain a specific distance from the blade and move at a controlled speed to avoid blur effects.

The time needed to complete both inspections may differ based on the wind turbine's size, which should be considered during the planning phase along with the vehicle's dynamical constraints. The~\acp{UAV} in this scenario are assumed to be multi-rotors, primarily quadrotors, with heterogeneous velocity and acceleration limitations. The objective is to plan trajectories for the~\acp{UAV} to efficiently complete the inspection mission while also satisfying the aforementioned constraints and maintaining a safe distance from obstacles and other~\acp{UAV} in the environment. It is assumed that a map of the environment, including a polyhedral representation of obstacles like the pylon and blades, is available prior to the inspection mission.



\section{Preliminaries}
\label{sec:preliminaries}

This section introduces the fundamental concepts required to grasp the contributions of this paper. We will provide a brief overview of the key system variables, \acl{STL}, and its robustness, smooth robustness approximations, the~\ac{STL} weighted generalization, and the \ac{STL} motion planning framework upon which the current contribution is built. To enhance readability, a summary of the notation is also provided in Table \ref{tab:tableOfNotation}.



\subsection{System definition}
\label{sec:systemDefinition}

Let us consider a discrete-time dynamical model of a drone represented in the general form $x_{k+1} = f(x_k, u_k)$, where $x_{k+1}, x_k \in \mathcal{X} \subset \mathbb{R}^n$ are the next and current states of the system, respectively, and $u_k \in \mathcal{U} \subset \mathbb{R}^m$ is the control input. Let us also assume $f \colon \mathcal{X} \times \mathcal{U} \rightarrow \mathcal{X}$ is differentiable in both arguments. Therefore, given an initial state $x_0 \in \mathcal{X}_0 \subset \mathbb{R}^n$ and a time vector $\mathbf{t} = (t_0, \dots, t_N)^\top \in \mathbb{R}^{N+1}$, with $N \in \mathbb{N}_{>0}$ being the number of samples that describe the evolution of the system with sampling period $T_s \in \mathbb{R}_{>0}$, we can define the finite control input sequence $\mathbf{u} = (u_0, \dots, u_{N-1})^\top \in \mathbb{R}^N$ as the input for the system to attain the unique sequence of states $\mathbf{x} = (x_0, \dots, x_{N})^\top \in \mathbb{R}^{N+1}$. Let us also introduce the notation $\mathcal{F}_W$ and $\mathcal{F}_B$ to denote the world frame and body frame reference systems, respectively.

Hence, we can define the state sequence $\mathbf{x}$ and control input sequence $\mathbf{u}$ for the $d$-th multi-rotor~\ac{UAV} as ${^d}\mathbf{x}=({^d}\mathbf{p}^{(1)}, {^d}\mathbf{v}^{(1)}, {^d}\mathbf{p}^{(2)}, {^d}\mathbf{v}^{(2)}, {^d}\mathbf{p}^{(3)}, {^d}\mathbf{v}^{(3)})^\top$ and ${^d}\mathbf{u}=({^d}\mathbf{a}^{(1)}, {^d}\mathbf{a}^{(2)}, {^d}\mathbf{a}^{(3)})^\top$, with ${^d}\mathbf{p}^{(j)}$, ${^d}\mathbf{v}^{(j)}$, and ${^d}\mathbf{a}^{(j)}$ representing vehicle's position, velocity, and acceleration sequences along the $j$-axis of the world frame $\mathcal{F}_W$, respectively, with $j=\{1,2,3\}$. Finally, the $k$-th elements of ${^d}\mathbf{p}^{(j)}$, ${^d}\mathbf{v}^{(j)}$, ${^d}\mathbf{a}^{(j)}$, and $\mathbf{t}$, are denoted as ${^d}p_k^{(j)}$, ${^d}v_k^{(j)}$, ${^d}a_k^{(j)}$, and $t_k$. 

\begin{table}
    \centering
    \caption{Notation--System variables, general symbols, and reference frames.}
    \vspace{-1.2em}
    \label{tab:tableOfNotation}
    \begin{tabular}{p{2.40cm} p{5.50cm}}
         \toprule
         $\mathcal{F}_W$, $\mathcal{F}_B$ & world and body reference frames \\
         $N$, $T_s$, $\mathbf{t}$, $t_k$, $\delta$ & number of samples, sampling period, time vector and its $k$-th element, members of the \ac{UAV} fleet\\
         ${^d}\mathbf{x}$, ${^d}\mathbf{u}$ & state and control input sequences of the $d$-th \ac{UAV} \\
         $M$, $AP$ & Set of real-valued functions and the corresponding predicates \\
         ${^d}\mathbf{p}$, ${^d}\mathbf{v}$, ${^d}\mathbf{a}$, ${^d}\psi$, ${^d}\bullet_k$ & position, velocity, acceleration, and heading of the $d$-th \ac{UAV} in $\mathcal{F}_W$ and their $k$-th elements\\
         ${^d}\mathbf{S}^{(j)}$ & splines encoding the drone motion primitives \\
         $\varphi$, $I$, $p_i$, $\mu_i$, $\lambda$ & \acs{STL} formula, generic time interval, $i$-th predicate and its real-valued function, tunable parameter for $\tilde{\rho}_\varphi(\mathbf{x})$ \\
         $\neg$, $\wedge$, $\vee$, $\hspace{-0.45em} \implies \hspace{-0.45em}$ & negation, conjunction, disjunction, and implication Boolean operators \\
         $\lozenge$, $\square$, $\bigcirc$, $\mathcal{U}$ & eventually, always, next and until temporal operators \\
         $\rho_\varphi(\mathbf{x}, t_k)$, $\tilde{\rho}_\varphi(\mathbf{x}, t_k)$ & robustness and smooth robustness values of the \ac{STL} formula $\varphi$ \\
         ${^d}\underline{v}^{(j)}$, ${^d}\underline{a}^{(j)}$, ${^d}\bar{v}^{(j)}$, ${^d}\bar{a}^{(j)}$ &  lower and upper limits for the velocity and acceleration of the $d$-th \ac{UAV} along the $j$-axis \\
         ${^d}\varphi_\mathrm{ws}$, ${^d}\varphi_\mathrm{obs}$, ${^d}\varphi_\mathrm{dis}$ & \ac{STL} safety requirements \\
         ${^d}\varphi_\mathrm{tr}$, ${^d}\varphi_\mathrm{bla}$ & \ac{STL} task requirements \\
         ${^d}\varphi_\mathrm{hm}$ & \ac{STL} mission completion requirement \\
         $T_N$, $T_\mathrm{ins}$, $T_\mathrm{bla}$ & mission duration, pylon and blade coverage time intervals \\
         $N_\mathrm{obs}$, $N_\mathrm{tr}$, $N_\mathrm{bla}$ & number of obstacles, number target areas, and number of blade sides to cover \\
         $\underline{p}^{(j)}_{\bullet}$, $\bar{p}^{(j)}_{\bullet}$ & generic vertices of the rectangular regions defining safety, task, and mission requirements \\
         $\Gamma_\mathrm{dis}$, $\Gamma_\mathrm{bla}$, $\varepsilon$, $\zeta$ & mutual distance threshold, minimum required distance away from the blade, maneuverability and safety margins \\ 
         $\mathcal{G}$, $\mathcal{V}$, $\mathcal{E}$, $\mathcal{W}$, $\mathcal{D}$, $\mathcal{T}$, $\mathcal{O}$ & graph, set of vertices, graph edges and weights,  set of drones, set of target areas and blade extreme points, and set of depots \\
         $\tau$, $o_\bullet$, $e_{ij}$, $w_{ij}$, $z_{ij}$ & cardinality of $\mathcal{T}$, generic element of $\mathcal{O}$, $\mathcal{E}$ and $\mathcal{W}$, integer variable for the \acs{MILP} solution \\ 
         $\mathcal{N}_i^\mathrm{in}$, $\mathcal{N}_i^\mathrm{out}$ & in-neighborhood and out-neighborhood sets of nodes \\
         ${^d}\hat{v}^{(j)}$, ${^d}\hat{a}^{(j)}$, ${^d}\check{v}^{(j)}$, ${^d}\check{a}^{(j)}$ & lower and upper limits for the revised velocity and acceleration of the $d$-th \ac{UAV} along the $j$-axis \\
         $\bar{\mathbf{t}}$, $E$, $\eta$, $\bar{t}_k$, $\hat{t}_k$ & event-trigger time vector and its dimension, event threshold value, generic entry of $\bar{\mathbf{t}}$, and time instance associated with the next task \\
         ${^\omega}\rho_\varphi(\bm{\omega}, \mathbf{x}, t_k)$, $\bm{\omega}$ & weighted version of the \ac{STL} robustness score and corresponding weight vector \\
         ${^d}T_c$, ${^d}\bm{\omega}_c$ & thrust and angular velocities of the $d$-th \ac{UAV} \\
         \bottomrule
    \end{tabular}
\end{table}

The model $x_{k+1} = f(x_k, u_k)$ considered for the drone is the one provided in~\cite{SilanoRAL2021}, where the motion primitives are encoded with splines, and here simply denoted, for each $j$-axis,  as $({^d}p_{k+1}^{(j)}, {^d}v_{k+1}^{(j)}, {^d}a_{k+1}^{(j)})^\top = {^d}\mathbf{S}^{(j)}({^d}p_{k}^{(j)}, {^d}v_{k}^{(j)}, {^d}a_{k}^{(j)})$. Next, we will use a label in the upper left to indicate a specific drone to which the dynamic model refers, while we will not use labels to indicate the vector stack of all drone variables.



\subsection{Signal temporal logic} 
\label{sec:signalTemporalLogic}

\ac{STL} concisely and unambiguously describes real-valued signal temporal behavior~\cite{Maler2004FTMA}. Unlike most used planning algorithms~\cite{LaValleBook},~\ac{STL} encapsulates all mission specifications in a single formula $\varphi$. For example, the statement ``visit regions A and B every $\SI{10}{\second}$, while always stay at least $\SI{1}{\meter}$ away from region C" can be expressed as a single~\ac{STL} formula, $\varphi$. The syntax and semantics of~\ac{STL} are detailed in~\cite{Maler2004FTMA, Donze2010FMATS}, but is omitted here for brevity.

In brief, an \ac{STL} formula $\varphi$ is constructed from a set of predicates $p_i$, where $i \in \mathbb{N}_0$, representing atomic prepositions like belonging to a region or comparisons of real values. Formally, let $M=\{\mu_1, \dots, \mu_L\}$ be a set of real-valued functions of the state, $\mu_i \colon \mathcal{X} \rightarrow \mathbb{R}$, and the corresponding set $AP\coloneqq \{p_1, \dots, p_L\}$ of predicates. Each predicate defines a set over the system state space, specifically $p_i$ defines $\{x \in \mathcal{X} \mid \mu_i(x) \geq 0\}$.

\ac{STL}'s grammar uses temporal operators, such as \textit{until} ($\mathcal{U}$), \textit{always} ($\square$), \textit{eventually} ($\lozenge$), and \textit{next} ($\bigcirc$), and logical operators like \textit{and} ($\wedge$), \textit{or} ($\vee$), \textit{negation} ($\neg$), and \textit{implication} ($\hspace{-0.35em} \implies \hspace{-0.35em}$), that act on atomic propositions over a non-singleton interval $I \subset \mathbb{R}$. Thus, an \ac{STL} formula $\varphi$ is built recursively from the predicates $p_i$ and using the grammar, as:
\begin{equation*}
    \varphi \coloneqq \top \vert p \vert \neg \varphi \vert \varphi_1 \vee \varphi_2 \vert \varphi_1 \wedge \varphi_2 \vert \square_I \varphi \vert \lozenge_I \varphi \vert \bigcirc_I \varphi \vert \varphi_1 \mathcal{U}_I \varphi_2,
\end{equation*}
where $\varphi_1$ and $\varphi_2$ are \ac{STL} formulae. These propositions are simple \textit{true} ($\top$) or \textit{false} ($\bot$) statements, such as belonging to a particular region or comparing real values. An~\ac{STL} formula $\varphi$ is valid if it evaluates to \textit{true} ($\top$) and invalid ($\bot$) otherwise. For example, $\varphi_1 \mathcal{U}_I \varphi_2$ requires that $\varphi_2$ holds within the time interval $I$ and that $\varphi_1$ holds uninterrupted until that point. 

If a formula $\varphi$ is \textit{time-bounded}, it contains no unbounded operator. The bound can be interpreted as the horizon of the future predicted system trajectory $\mathbf{x}$ that is needed to calculate the satisfaction of $\varphi$. Generally, to evaluate whether such a formula $\varphi$ holds on a given trajectory, only a finite-length prefix of that trajectory is needed~\cite{BeltaARC2019, Belta2017BBook}. In this paper, we will only consider bounded time intervals $I$.



\subsection{Robust signal temporal logic}
\label{sec:robustSignalTemporalLogic}

Uncertainties and unforeseen events can affect the satisfaction of an~\ac{STL} formula $\varphi$ (see Section~\ref{sec:signalTemporalLogic}). \textit{Robust semantics} for~\ac{STL} formulae~\cite{Maler2004FTMA, Donze2010FMATS, Fainekos2009TCS} account for these factors by ensuring a margin of satisfaction, measuring how well (poorly) a given specification is satisfied. The \textit{robustness} metric, $\rho$, guides the optimization process in finding the best feasible solution to meet mission requirements. It can be formally defined using the following recursive formulae: 
\begin{equation*}
    \begin{array}{rll}
    \rho_{p_i} (\mathbf{x}, t_k) & = & \mu_i (x(t_k)), \\ 
    \rho_{\neg \varphi} (\mathbf{x}, t_k) & = & - \rho_\varphi (\mathbf{x}, t_k), 
    \\
    \rho_{\varphi_1 \wedge \varphi_2} (\mathbf{x}, t_k) & = & \min \left(\rho_{\varphi_1} (\mathbf{x}, t_k), \rho_{\varphi_2} (\mathbf{x}, t_k) \right), \\
    \rho_{\varphi_1 \vee \varphi_2} (\mathbf{x}, t_k) & = & \max \left(\rho_{\varphi_1} (\mathbf{x}, t_k), \rho_{\varphi_2} (\mathbf{x}, t_k) \right), \\
    \rho_{\square_I \varphi} (\mathbf{x}, t_k) & = & \min\limits_{t_k^\prime \in [t_k + I]} \rho_\varphi (\mathbf{x}, t_k^\prime), \\
    \rho_{\lozenge_I \varphi} (\mathbf{x}, t_k) & = & \max\limits_{t_k^\prime \in [t_k + I]} \rho_\varphi (\mathbf{x}, t_k^\prime), \\
    \rho_{\bigcirc_I \varphi} (\mathbf{x}, t_k) & = & \rho_\varphi (\mathbf{x}, t_k^\prime), \text{with} \; t_k^\prime \in [t_k+I], \\
    \rho_{\varphi_1 \mathcal{U}_I \varphi_2} (\mathbf{x}, t_k) & = & \max\limits_{t_k^\prime \in [t_k + I]} \Bigl( \min \left( \rho_{\varphi_2} (\mathbf{x}, t_k^\prime) \right), \\
    & &  \hfill \min\limits_{ t_k^{\prime\prime} \in [t_k, t_k^\prime] } \left( \rho_{\varphi_1} (\mathbf{x}, t_k^{\prime \prime} \right)  \Bigr),
    \end{array}
\end{equation*}
where $t_k + I$ represents the Minkowski sum of scalar $t_k$ and time interval $I$. These formulae, as said, are recursively defined from \textit{predicates} $p_i$ and their corresponding real-valued function $\mu_i(x(t_k))$, which are true if their robustness value is greater than zero and false otherwise. 

The overall behavior of the formula is logical, becoming false if any predicates within it are false. In simpler terms, the \ac{STL} formula $\varphi_1 \vee \varphi_2$ is satisfied if $\varphi_1$ or $\varphi_2$ is true. Evaluation proceeds by applying logical and temporal operators (such as \textit{always}, \textit{eventually}, \textit{conjunction}, etc.) from the innermost to the outermost part of the formula. For example, this could entail conditions like being inside a target region or outside an obstacle region, each defined by a specific set of predicates. More comprehensive explanations can be found in references~\cite{Maler2004FTMA, Donze2010FMATS, Fainekos2009TCS}. In this context, we say that $\mathbf{x}$ satisfies the~\ac{STL} formula $\varphi$ at time $t_k$ (shortened as $\mathbf{x}(t_k) \models \varphi$) if $\rho_\varphi(\mathbf{x}, t_k) > 0$, and violates it if $\rho_\varphi(\mathbf{x}, t_k) \leq 0$. Furthermore, the value of  $\rho_\varphi(\mathbf{x}, t_k)$ represents ``how well'' the formula is satisfied (if $\rho_\varphi(\mathbf{x}, t_k)>0$) or ``how much'' is violated (if $\rho_\varphi(\mathbf{x}, t_k)\leq 0$), implicitly introducing a robustness criterion. 

Hence, we compute control inputs $\mathbf{u}$, maximizing robustness $\rho_\varphi(\mathbf{x}, t_k)$ over finite state $\mathbf{x}$ and input sequences $\mathbf{u}$, with $\mathbf{u}$ and $\mathbf{x}$ obeying to system dynamics. The optimal sequence for input and states is denoted as $\mathbf{u}^\star$ and  $\mathbf{x}^\star$. A larger $\rho_\varphi (\mathbf{x}^\star, t_k)$ signifies a more resilient system behavior against disturbances, allowing the system to withstand greater values of the latter without violating the \ac{STL} specification.
To simplify notation, we will use $\rho_\varphi(\mathbf{x})$ instead of $\rho_\varphi(\mathbf{x}, 0)$ when $t_k = 0$.



\subsection{Smooth approximation}
\label{sec:smoothApproximation}

The calculation of $\rho_\varphi(\mathbf{x})$ incorporates non-differentiable functions like $\min$ and $\max$. To tackle the computational challenges linked with these non-differentiable functions, it is advantageous to utilize a smooth approximation, represented as $\tilde{\rho}_\varphi(\mathbf{x})$, of the robustness function $\rho_\varphi(\mathbf{x})$. This smoothed approximation offers a more manageable and computationally efficient solution. Hence,
let $\lambda \in \mathbb{R}_{>0}$ be a tunable parameter. The smooth approximation of the $\min$ and $\max$ operators with $\beta$ predicate arguments is~\cite{Gilpin2021LCSS}:
\begin{equation*}
    \begin{split}
    &\max(\rho_{\varphi_1}, \dots, \rho_{\varphi_\beta}) \approx \frac{ \sum_{i=1}^\beta  \rho_{\varphi_i} e^{\lambda \rho_{\varphi_i}} }{ \sum_{i=1}^\beta  e^{\lambda \rho_{\varphi_i}} }, 
    \\
    &\min(\rho_{\varphi_1}, \dots, \rho_{\varphi_\beta}) \approx -\frac{1}{\lambda} \log \left( \sum_{i=1}^\beta 
    e^{-\lambda \rho_{\varphi_i}} \right). 
    \end{split}
\end{equation*}

Compared to our prior work~\cite{SilanoRAL2021}, we adopt an \textit{asymptotically complete} and \textit{smooth everywhere} approximation, which resembles the widely recognized~\acl{LSE} approximation~\cite{Donze2010FMATS}. The proposed approximation does not overestimate the $\max$ operator and, as a result, is sound. \textit{Soundness} indicates that a suitable sequence $\mathbf{u}^\star$ with strictly positive smooth robustness ($\tilde{\rho}_\varphi(\mathbf{x}) > 0$) satisfies the specification $\varphi$, whereas a sequence $\mathbf{u}^\star$ with strictly negative smooth robustness ($\tilde{\rho}_\varphi(\mathbf{x}) < 0$) violates it. The term \textit{asymptotical completeness} implies that the resulting approximation $\tilde{\rho}_\varphi(\mathbf{x})$ for the final robustness formula can approach the true robustness $\rho_\varphi(\mathbf{x})$ arbitrarily closely as $\lambda$ tends towards infinity ($\lambda \rightarrow \infty$). Furthermore, the proposed approximation is \textit{smooth everywhere}, possessing infinite differentiability, thus making it viable to use gradient-based optimization algorithms to ascertain a solution~\cite{Gilpin2021LCSS}. By increasing $\lambda$, the approximation can better reflect the true robustness (see Section~\ref{sec:robustSignalTemporalLogic}).



\subsection{Weighted signal temporal logic}
\label{sec:weightedSignalTemporalLogic}

In many applications, high-level temporal logic specifications may include obligatory or alternative sub-specifications or timings with varying importance or priorities. Traditional \ac{STL} lacks the expressivity to specify these preferences. For example, consider $\varphi = \lozenge_I (\mathbf{x} > 0)$, which is satisfied if $\mathbf{x}$ becomes greater than $0$ within the time interval $I$. However, satisfaction at earlier times within this deadline may be more desirable. 

Assisting importance and priorities becomes crucial, especially when a formula contains conflicting obligatory sub-formulae. Therefore, an extension of~\ac{STL}, known as \ac{wSTL}~\cite{MehdipourLCSS2021, Cardona2023ACM, CardonaECC2023}, can be employed to enable user preferences such as priorities and importance. The syntax of \ac{wSTL} is an extension of the \ac{STL} syntax, and is defined as:
\begin{equation*}
    \varphi \coloneqq \top \vert p \vert \neg \varphi \vert \bigwedge_{i=1}^{N} {^\omega}\varphi_i \vert \bigvee_{i=1}^{N} {^\omega}\varphi_i \vert \square_I {^\omega}\varphi \vert \lozenge_I {^\omega}\varphi \vert \bigcirc_I {^\omega}\varphi \vert {^\omega}\varphi_1 \mathcal{U}_I {^\omega}\varphi_2,
\end{equation*}
where the logical \textit{true} ($\top$) and \textit{false} ($\bot$) values, the predicate $p$, and all the Boolean and temporal operators have the same interpretation as in \ac{STL} (as discussed in Section~\ref{sec:signalTemporalLogic}). The weight $\bm{\omega} = [\omega_i]_{i=1}^N \in \mathbb{R}_{>0}^N$ assigns a positive weight $\omega_i$ to each sub-formula $i$ of the $N$ sub-formulae of the Boolean operators, and $\bm{\omega} = [\omega_{t_k}]_{t_k \in I} \in \mathbb{R}_{>0}^{|I|}$ assigns a positive weight $\omega_{t_k}$ to time $t_k$ in the interval $I$ of the temporal operators and $|I|$ denotes cardinality of $I$. 

The weights $\omega$ capture the \textit{importance} of obligatory specifications for conjunctions ($\wedge$) or \textit{priorities} of alternatives for disjunctions ($\vee$). Similarly, $\omega$ captures the \textit{importance} of satisfaction times for temporal \textit{always} ($\square$) or \textit{priorities} of satisfaction times for temporal \textit{eventually} ($\lozenge$) over the interval $I$. Throughout the paper, if the weight $\omega$ associated with an operator (Boolean or temporal) in a \ac{wSTL} formula ${^\omega}\varphi$ is constant $1$, we drop it from the notation. Thus, \ac{STL} formulae are \ac{wSTL} formulae with all weights equal to $1$.

Given a \ac{wSTL} specification ${^\omega}\varphi$, the weighted robustness score ${^\omega}\rho_\varphi(\bm{\omega}, \mathbf{x}, t_k)$ is recursively defined as:
\begin{equation*}
    \begin{array}{rll}
    {^\omega}\rho_{p_i} (\bm{\omega}, \mathbf{x}, t_k) & = & \mu_i (\bm{\omega}, x(t_k)), \\ 
    {^\omega}\rho_{\neg \varphi} (\bm{\omega}, \mathbf{x}, t_k) & = & - \rho_\varphi (\bm{\omega}, \mathbf{x}, t_k), 
    \\
    {^\omega}\rho_{\varphi_1 \wedge \varphi_2} (\bm{\omega}, \mathbf{x}, t_k) & = & \min \left(\rho_{\varphi_1} (\bm{\omega}, \mathbf{x}, t_k), \rho_{\varphi_2} (\bm{\omega}, \mathbf{x}, t_k) \right), \\
    {^\omega}\rho_{\varphi_1 \vee \varphi_2} (\bm{\omega}, \mathbf{x}, t_k) & = & \max \left(\rho_{\varphi_1} (\bm{\omega}, \mathbf{x}, t_k), \rho_{\varphi_2} (\bm{\omega}, \mathbf{x}, t_k) \right), \\
    {^\omega}\rho_{\square_I \varphi} (\bm{\omega}, \mathbf{x}, t_k) & = & \min\limits_{t_k^\prime \in [t_k + I]} \rho_\varphi (\bm{\omega}, \mathbf{x}, t_k^\prime), \\
    {^\omega}\rho_{\lozenge_I \varphi} (\bm{\omega}, \mathbf{x}, t_k) & = & \max\limits_{t_k^\prime \in [t_k + I]} \rho_\varphi (\bm{\omega}, \mathbf{x}, t_k^\prime), \\
    {^\omega}\rho_{\bigcirc_I \varphi} (\bm{\omega}, \mathbf{x}, t_k) & = & \rho_\varphi (\bm{\omega}, \mathbf{x}, t_k^\prime), \text{with} \; t_k^\prime \in [t_k+I], \\
    {^\omega}\rho_{\varphi_1 \mathcal{U}_I \varphi_2} (\bm{\omega}, \mathbf{x}, t_k) & = & \max\limits_{t_k^\prime \in [t_k + I]} \Bigl( \min \left( \rho_{\varphi_2} (\bm{\omega}, \mathbf{x}, t_k^\prime) \right), \\
    & &  \hfill \min\limits_{ t_k^{\prime\prime} \in [t_k, t_k^\prime] } \left( \rho_{\varphi_1} (\bm{\omega}, \mathbf{x}, t_k^{\prime \prime} \right)  \Bigr),
    \end{array}
\end{equation*}
where, as discussed for \ac{STL} (see Section~\ref{sec:signalTemporalLogic}), $t_k + I$ represents the Minkowski sum of scalar $t_k$ and time interval $I$. These formulae consist of predicates, $p_i$, along with their corresponding weighted real-valued function $\mu_i(\bm{\omega}, \mathbf{x}, t_k)$, which are true if their robustness value is greater than zero and false otherwise.



\subsection{\ac{STL} motion planner}
\label{sec:STLMotionPlanner}

By encoding the mission specifications detailed in Section~\ref{sec:problemDescription} as an~\ac{STL} formula $\varphi$ and replacing its robustness $\rho_\varphi(\mathbf{x}, t_k)$ with the smooth approximation $\tilde{\rho}_\varphi(\mathbf{x}, t_k)$ (see Section~\ref{sec:smoothApproximation}), the problem of generating trajectories for the multi-rotors can be formulated as~\cite{SilanoRAL2021}:
\begin{equation}\label{eq:optimizationProblemMotionPrimitives}
    \begin{split}
    &\maximize_{{^d}\mathbf{p}^{(j)}, \, {^d}\mathbf{v}^{(j)}, \,{^d}\mathbf{a}^{(j)}\atop d \in \mathcal{D}} \;\;
    {\tilde{\rho}_\varphi (\mathbf{p}^{(j)}, \mathbf{v}^{(j)} )} \\
    &\qquad \,\;\, \text{s.t.}~\quad\, {^d}\underline{v}^{(j)} \leq {^d}v^{(j)}_k \leq {^d}\bar{v}^{(j)}, \\
    &\,\;\;\;\, \qquad \qquad {^d}\underline{a}^{(j)} \leq {^d}a^{(j)}_k \leq {^d}\bar{a}^{(j)}, \\
    &\,\;\;\;\, \qquad \qquad \tilde{\rho}_\varphi ({^d}\mathbf{p}^{(j)}, \, {^d}\mathbf{v}^{(j)}) \geq \zeta, \\
    &\,\;\;\;\, \qquad \qquad {^d}\mathbf{S}^{(j)}, \forall k=\{0,1, \dots, N-1\}
    \end{split},
\end{equation}
where ${^d}\underline{v}^{(j)}$ and ${^d}\underline{a}^{(j)}$ represent the lower limits of velocity and acceleration, respectively, while ${^d}\bar{v}^{(j)}$ and ${^d}\bar{a}^{(j)}$ denote their respective upper limits, for drone $d$ along each $j$-axis of the world frame $\mathcal{F}_W$. Here, $\mathcal{D}$ denotes the set of drones, and $\mathbf{p}^{(j)}$ and $\mathbf{v}^{(j)}$ concatenate the position and velocity of all drones. The minimum robustness threshold, denoted as $\tilde{\rho}_\varphi (\mathbf{p}^{(j)}, \mathbf{v}^{(j)}) \geq \zeta$, acts as a safety buffer for ensuring the satisfaction of the~\ac{STL} formula $\varphi$ even in the presence of disturbances. As illustrated in~\cite{PantCDC2017}, disturbances below magnitude $\zeta$ do not result in formula violations. The specific value of $\zeta$ can be determined such that $\lvert \rho_\varphi(\mathbf{x}) - \tilde{\rho}_\varphi(\mathbf{x}) \rvert \leq \zeta$. Moreover, the shorthand notation ${^d}\mathbf{S}^{(j)}$ refers to the motion primitives satisfying the dynamics of drone $d$ along each $j$-axis, as explained in Section~\ref{sec:systemDefinition}. For simplicity, we assume symmetric velocity and acceleration limits.



\section{Problem Solution}
\label{sec:problemSolution}

In this section, we introduce a method to generate trajectories for a fleet of $\delta$ multi-rotor~\acp{UAV} belonging to set $\mathcal{D}$, i.e., $\delta = \lvert \mathcal{D} \lvert$. The mission specifications have been described in Section~\ref{sec:problemDescription} and are expressed as an~\ac{STL} formula $\varphi$ (Section~\ref{sec:specificationMapping}). The motion planner is a nonlinear, non-convex max-min optimization problem  that accounts for mission specifications and physical constraints. However, finding optimal solutions within a reasonable time frame can be challenging because solvers easily get stuck in local optima depending on the initial guess~\cite{Bertsekas2009Book}. To overcome this challenge, we propose a two-step hierarchical approach that simplifies the mission into an~\ac{MILP} planning problem (Section~\ref{sec:milpEncoding}) before seeding the global~\ac{STL} optimizer (see Section~\ref{sec:STLMotionPlanner}). Our framework also includes an event-triggered replanner to cope with disturbances or unforeseen events (Section~\ref{sec:eventBasedReplanner}), as well as user preferences to handle potential issues arising from conflicting tasks (Section~\ref{sec:attrititionAwarePlanner}).



\subsection{Specification mapping}
\label{sec:specificationMapping}

This section aims to design the mission specifications for the problem described in Section~\ref{sec:problemDescription} and derive the corresponding~\ac{STL} formula, $\varphi$. While a wind turbine inspection could theoretically be performed safely with just one~\ac{UAV}, practical considerations necessitate cooperative execution by multiple~\acp{UAV}. Factors such as efficiency, coverage area, time-saving, enhanced safety, and redundancy in case of unexpected events can benefit from the involvement of multiple drones to ensure mission success. Therefore, the wind turbine inspection is encoded as a cooperative execution by the~\acp{UAV} to meet safety and task requirements. 

The \textit{safety requirements} must be met throughout the entire operation time $T_N$. Hence, the~\acp{UAV} must adhere to three specifications: remaining within the workspace (${^d}\varphi_\mathrm{ws}$), avoiding obstacles to prevent collisions (${^d}\varphi_\mathrm{obs}$), and keeping a safe distance from other~\acp{UAV} (${^d}\varphi_\mathrm{dis}$). The \textit{task requirements} include pylon and blade inspections. To inspect the pylon (${^d}\varphi_\mathrm{tr}$), the~\ac{UAV} must visit all target areas once and remain there for at least an inspection time $T_\mathrm{ins}$. In order to inspect the blade (${^d}\varphi_\mathrm{bla}$), the~\ac{UAV} must approach the leading edge, maintain a certain distance from the blade surface while covering it at a limited speed, and reach the rotor shaft before turning to inspect the other side of the blade surface. 
The coverage task must last at least $T_\mathrm{bla}$.

Notably, for the case scenario, we assume that each~\ac{UAV} can only cover some blades or targets, reflecting a scenario where some drones may be equipped with specialized instruments or sensors, resulting in their heterogeneous physical capabilities. This assumption necessitates coordination among multiple~\acp{UAV} to ensure completion of all required inspections. Here, we do not specify which task (pylon or blade inspection) a single drone must accomplish. Instead, we allow the framework to choose the most convenient option, while also leaving open the possibility of accomplishing both tasks.
Finally, each~\ac{UAV} must return to its initial position after completing its inspection operations (${^d}\varphi_\mathrm{hm}$) and remain there thereafter. 

All the above mission specifications can be represented in the~\ac{STL} formula:
\begin{equation}\label{eq:windTurbineInspectionFormula}
    \begin{split}
    \varphi =&  \bigwedge_{d\in\mathcal{D}}\square_{[0,T_N]} ( {^d}\varphi_{\mathrm{ws}}\wedge {^d}\varphi_{\mathrm{obs}}\wedge {^d}\varphi_{\mathrm{dis}} ) \,
    \wedge \\
    & \bigwedge_{q=1}^{N_\mathrm{tr}}\lozenge_{[0,T_N-T_{\mathrm{ins}}]} \bigvee_{d\in\mathcal{D}}\square_{[0,T_{\mathrm{ins}}]} {^d}\varphi_{\mathrm{tr,q}} \,
    \wedge \\
    & \bigwedge_{q=1}^{N_\mathrm{bla}}\lozenge_{[0,T_N-T_{\mathrm{bla}}]} \bigvee_{d\in\mathcal{D}}\square_{[0,T_{\mathrm{bla}}]} {^d}\varphi_{\mathrm{bla,q}} \, 
    \wedge \\
    & \bigwedge_{d\in\mathcal{D}} \lozenge_{[1,T_N]}{^d}\varphi_{\mathrm{hm}}\, 
    \wedge \\
    & \bigwedge_{d\in\mathcal{D}} \square_{[1,T_N-1]}\left( {^d}\varphi_{\mathrm{hm}} \hspace{-0.45em} \implies \hspace{-0.45em} \bigcirc_{[0, t_k + 1]} {^d}\varphi_{\mathrm{hm}} \right).
   \end{split}
\end{equation}

\begin{figure}[tb]
    \begin{center}
    \vspace*{-2em}
    \hspace*{10mm}
        \adjincludegraphics[trim={{.21\width} {.05\height} {0.0\width}	{.0\height}},clip,scale=0.25]{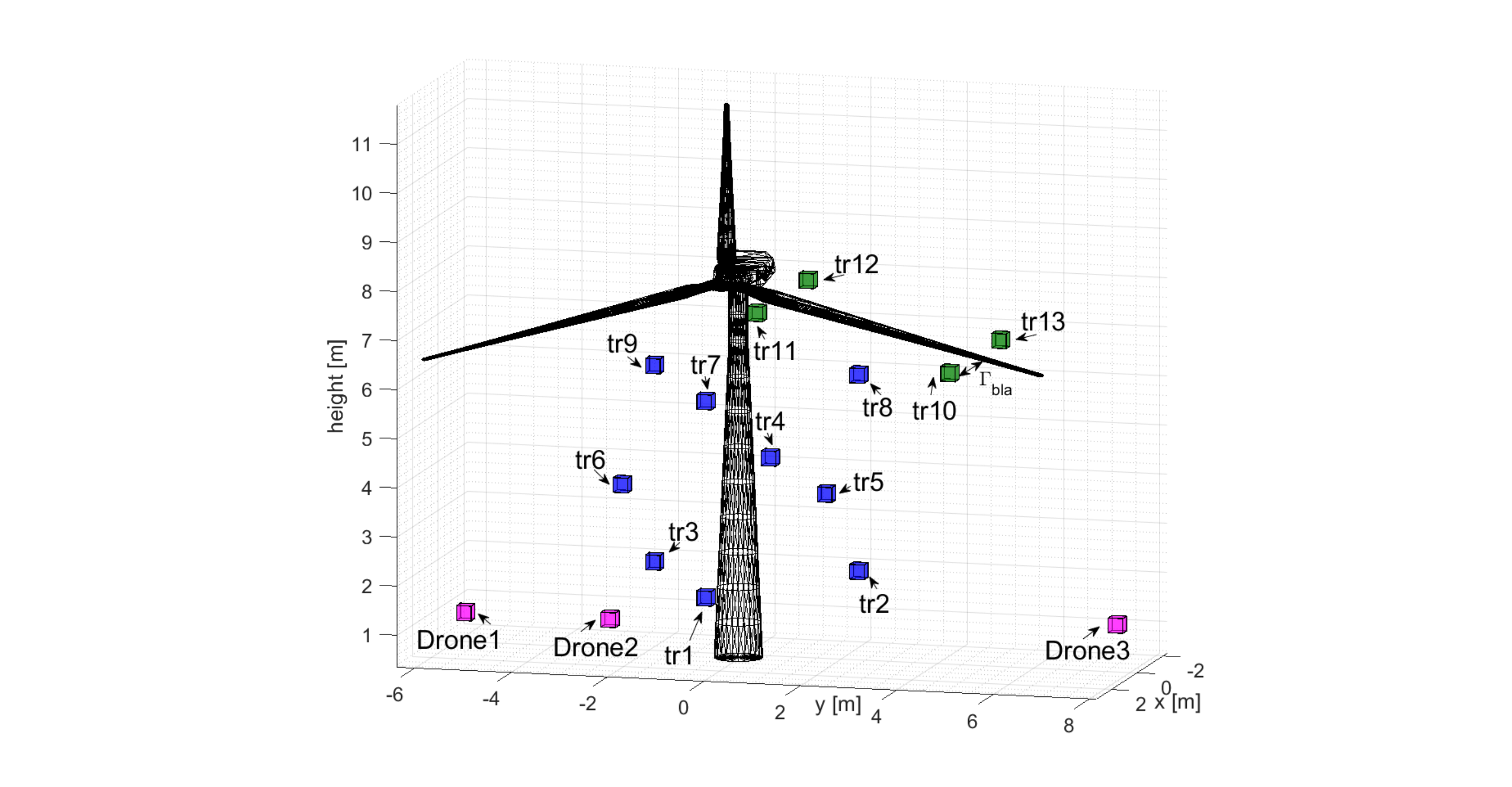}
	\vspace*{-10mm}
    \end{center}
    \caption{Wind turbine inspection scenario. Target areas and blade extreme points are represented in blue and green, respectively, while the drones' starting points are in magenta.}
    \label{fig:inspectionScenarioNoTraj}
\end{figure}

Here, the~\ac{STL} formula $\varphi$ comprises six specifications (${^d}\varphi_\mathrm{ws}$, ${^d}\varphi_\mathrm{obs}$, ${^d}\varphi_\mathrm{dis}$, ${^d}\varphi_\mathrm{tr}$, ${^d}\varphi_\mathrm{bla}$, and ${^d}\varphi_\mathrm{hm}$) and three time intervals ($T_N$, $T_\mathrm{ins}$, and $T_\mathrm{bla}$). 
The following equations describe each of these specifications: 
\begin{subequations}\label{eq:STLcomponents}
    \begin{align}
    \textstyle{ {^d}\varphi_\mathrm{ws}} &= \textstyle{ \bigwedge_{j=1}^3 {^d}\mathbf{p}^{(j)} \hspace{-0.25em} \in (\underline{p}^{(j)}_\mathrm{ws}, \bar{p}^{(j)}_\mathrm{ws})}, \label{subeq:belongWorkspace} \\
    \textstyle{ {^d}\varphi_\mathrm{obs}} &= \textstyle{ \bigwedge_{j=1}^3\bigwedge_{q=1}^{N_\mathrm{obs}} {^d}\mathbf{p}^{(j)} \hspace{-0.25em} \not\in (\underline{p}_{\mathrm{obs,q}}^{(j)}, \bar{p}_{\mathrm{obs,q}}^{(j)})}, \label{subeq:avoidObostacles} \\
    \textstyle{{^d}\varphi_\mathrm{hm}} &= \textstyle{\bigwedge_{j=1}^3 {^d}\mathbf{p}^{(j)} \hspace{-0.25em} \in (\underline{p}^{(j)}_\mathrm{hm}, \bar{p}^{(j)}_\mathrm{hm})}, \label{subeq:backHome} \\
    \textstyle{{^d}\varphi_\mathrm{dis}} &=  \textstyle{\bigwedge_{ \{d, m\} \in \mathcal{D}, d \neq m } \hspace{0.2em} \lVert {^d}\mathbf{p} - {^m}\mathbf{p} \rVert \geq \Gamma_\mathrm{dis}}, \label{subeq:keepDistance} \\
    %
    %
    \textstyle{{^d}\varphi_{\mathrm{tr,q}}} &=  \textstyle{\bigwedge_{j=1}^3  {^d}\mathbf{p}^{(j)} \hspace{-0.25em} \in (\underline{p}^{(j)}_{\mathrm{tr,q}}, \bar{p}^{(j)}_{\mathrm{tr,q}})}, \label{subeq:visitTargets} 
\end{align}
\end{subequations}
\begin{subequations}
    \begin{align}
    \textstyle{{^d}\varphi_\mathrm{bla,q}} &= \textstyle{\bigwedge_{j=1}^3 {^d}\mathbf{p}^{(j)} \hspace{-0.25em} \in (\underline{p}^{(j)}_{\mathrm{bla,q}}, \bar{p}^{(j)}_{\mathrm{bla,q}}) \; \wedge  \nonumber} \\ 
     & \hspace{1em} \textstyle{\mathrm{dist}_{\mathrm{bla,q}}({^d}\mathbf{p})\in (\Gamma_{\mathrm{bla}}-\varepsilon, \Gamma_{\mathrm{bla}}+\varepsilon)}, \label{subeq:bladeInspection} \tag{3f} 
\end{align}
\end{subequations} \setcounter{equation}{3}
where~\eqref{subeq:belongWorkspace} constrains the position of each~\ac{UAV} to remain within the designated workspace, with $\underline{p}^{(j)}_\mathrm{ws}$ and $\bar{p}^{(j)}_\mathrm{ws}$ denoting the working space limits. Obstacle avoidance and mission completion operation are defined in~\eqref{subeq:avoidObostacles} and~\eqref{subeq:backHome}, respectively, where $N_\mathrm{obs}$ denotes the number of obstacles in the map. Rectangular regions with vertices denoted by $\underline{p}_\mathrm{obs,q}^{(j)}$, $\underline{p}_\mathrm{hm}^{(j)}$, $\bar{p}_\mathrm{obs,q}^{(j)}$, and $\bar{p}_\mathrm{hm}^{(j)}$ define obstacle and drone's initial position areas.~\eqref{subeq:keepDistance} encodes the safety distance requirement, where $\Gamma_\mathrm{dis} \in \mathbb{R}_{>0}$ represents a threshold value for the drones mutual distance. Finally, inspecting the pylon and blade is done with~\eqref{subeq:visitTargets} and~\eqref{subeq:bladeInspection}, respectively, where $N_\mathrm{tr}$ and $N_\mathrm{bla}$ indicate the number of target areas and blade sides to cover, respectively. The vertices  $\underline{p}_\mathrm{tr,q}^{(j)}$, $\underline{p}_\mathrm{bla,q}^{(j)}$, $\bar{p}_\mathrm{tr,q}^{(j)}$, and $\bar{p}_\mathrm{bla,q}^{(j)}$ identify the target and covering areas, with the leading edge and rotor shaft serving as the extreme points for each side of the blade surface. The function $\mathrm{dist}_{\mathrm{bla,q}}(\cdot)$ in~\eqref{subeq:bladeInspection} calculates the Euclidean distance between the position of the \ac{UAV} (${^d}\mathbf{p}$) and the corresponding point on blade $q$, determined by projecting the~\ac{UAV}'s position orthogonally onto the surface of the blade. Its role in~\eqref{subeq:bladeInspection} is to enforce a certain distance between the~\ac{UAV} and the blade surface, with $\Gamma_\mathrm{bla} \in \mathbb{R}_{>0}$ and $\varepsilon\in \mathbb{R}_{>0}$ representing the minimum required distance and maneuverability margin, respectively. The latter enables the~\ac{UAV} to continue operating under challenging scenarios, such as collision risks, while meeting mission specifications. The \textit{always} operators ($\square$) are exploited to ensure the satisfaction of the minimum time requirements $T_\mathrm{bla}$ and $T_\mathrm{ins}$. Figure~\ref{fig:inspectionScenarioNoTraj} illustrates the scenario.

Utilizing the specifications outlined in~\eqref{eq:windTurbineInspectionFormula}, the optimization problem, as defined in Section~\ref{sec:STLMotionPlanner}, is formulated to determine feasible trajectories that maximize the smooth robustness $\tilde{\rho}_\varphi(\mathbf{x})$ concerning the given mission specifications $\varphi$. To achieve this objective, it is essential to compute the robustness score for each predicate. The~\ac{STL} formula~\eqref{eq:windTurbineInspectionFormula} consists of two types of predicates. The first type assesses whether the~\acp{UAV}' position belongs or does not belong to a specific region, as illustrated in~\eqref{subeq:belongWorkspace},~\eqref{subeq:avoidObostacles},~\eqref{subeq:backHome},~\eqref{subeq:visitTargets},~\eqref{subeq:bladeInspection}. The second type evaluates the distance between~\acp{UAV}, as indicated by the safety requirement in~\eqref{subeq:keepDistance}. The robustness values are determined based on the Euclidean distance. For predicates belonging to the first group, a positive robustness signifies that the~\ac{UAV} lies within the designated region. The robustness increases as the minimum Euclidean distance to the boundaries of the region along the trajectory grows larger. However, in the case of~\eqref{subeq:avoidObostacles}, the opposite holds true, where being within the obstacle region corresponds to a negative robustness. In the safety distance predicate~\eqref{subeq:keepDistance}, the robustness is positive when the distance between~\acp{UAV} exceeds the threshold $\Gamma_\mathrm{dis}$. The robustness value increases as the minimum Euclidean distance between~\acp{UAV} along the trajectory becomes larger.
 


\subsection{MILP encoding}
\label{sec:milpEncoding}

An appropriate initial guess is crucial for obtaining optimal solutions for the~\ac{STL} motion planner within a reasonable time frame and avoiding the solver from getting trapped in local optima. To achieve this, the original wind turbine inspection problem is abstracted to create a simpler optimization problem with fewer constraints. The initial guess considers the mission requirements related to covering the blade surface and visiting all of the target areas (${^d}\varphi_\mathrm{bla}$ and ${^d}\varphi_\mathrm{tr}$). However, the constraints related to obstacle avoidance, workspace, safety distance, and mission completion requirements (${^d}\varphi_\mathrm{obs}$, ${^d}\varphi_\mathrm{ws}$, ${^d}\varphi_\mathrm{dis}$, and ${^d}\varphi_\mathrm{hm}$) and mission time intervals ($T_N$, $T_\mathrm{ins}$, and $T_\mathrm{bla}$) are ignored to reduce complexity in the problem.
To formulate the~\ac{MILP} planning problem, a graph-based representation is employed to connect the target areas, blade extreme points (leading edge and rotor shaft), and the vehicles' initial positions. To guarantee seamless coverage, the four blade extreme points (two on each side) are considered as a single node in the graph. The~\ac{MILP} assigns tasks to the vehicles and determines a navigation sequence for each~\ac{UAV}.  
This results in a type of~\ac{VRP}~\cite{LaValleBook}, which assigns tasks to vehicles and provides a navigation sequence for each~\ac{UAV}. 
Furthermore, we included the mission time in the optimization to distribute it equally among fleet members.

\begin{figure}[tb]
    \vspace{-2.25em}
    \scalebox{1.05}{
    \begin{tikzpicture}[node distance=17.5mm, on grid, auto] 
        \node[draw, circle, double] (1) {\scriptsize{$0_{|1}$}};  
        \node[draw, circle, solid]  (2) [above right of=1] {\scriptsize{$1$}};  
        \node[draw, circle, dashed]  (3) [above of=2] {\scriptsize{$4$}};  
        \node[draw, circle, double] (4) [below right of=2] {\scriptsize{$0_{|2}$}}; 
        \node[draw, circle, solid]  (5) [above right of=4] {\scriptsize{$3$}};  
        \node[draw, circle, solid] (6) [above left of=1] {\scriptsize{$2$}};  
        
        \draw[-] (1) to [out=50, in=235, looseness=0] node[midway, above, sloped, pos=0.5, yshift=-9.5pt] {\scalebox{0.6}{$w_{01|1}$}} (2);
        \draw[-] (2) to [out=-35, in=215, looseness=0.45] node[midway, above, sloped, pos=0.5, yshift=-9.5pt] {\scalebox{0.6}{$w_{13|1}$}} (5);
        \draw[-] (1) to [out=180, in=160, looseness=2.0] node[midway, above, sloped, pos=0.5, yshift=-3.25pt] {\scalebox{0.6}{$w_{04|1}$}} (3);
        \draw[-] (1) to [out=-65, in=-60, looseness=1.10] node[midway, above, sloped, pos=0.55, yshift=-9.5pt] {\scalebox{0.6}{$w_{03|1}$}} (5);
        \draw[-] (6) to [out=35, in=145, looseness=0.45] node[midway, above, sloped, pos=0.5, yshift=-3.5pt] {\scalebox{0.6}{$w_{12|1}$}} (2);
        \draw[-] (1) to [out=145, in=-55, looseness=0] node[midway, above, sloped, pos=0.5, yshift=-3.75pt] {\scalebox{0.6}{$w_{02|1}$}} (6);
        \draw[-] (3) to [out=315, in=45, looseness=0.45] node[midway, above, sloped, pos=0.5, yshift=-3.25pt] {\scalebox{0.6}{$w_{41|1}$}} (2);
        \draw[-] (3) to [out=200, in=55, looseness=0.35] node[midway, above, sloped, pos=0.5, yshift=-3.25pt] {\scalebox{0.6}{$w_{42|1}$}} (6);
        \draw[-] (5) to [out=120, in=-17, looseness=0.65] node[midway, above, sloped, pos=0.5, yshift=-9pt] {\scalebox{0.6}{$w_{34|1}$}} (3);
        \draw[-] (5) to [out=270, in=265, looseness=1.22] node[midway, above, sloped, pos=0.90, yshift=-4.75pt] {\scalebox{0.6}{$w_{32|1}$}} (6);
			
        \draw[-] (5) to [out=145, in=35, looseness=0.45] node[midway, above, sloped, pos=0.5, yshift=-3.75pt] {\scalebox{0.6}{$w_{13|2}$}} (2);
        \draw[-] (4) to [out=140, in=-55, looseness=0] node[midway, above, sloped, pos=0.5, yshift=-9.5pt] {\scalebox{0.6}{$w_{01|2}$}} (2);
        \draw[-] (4) to [out=20, in=-110, looseness=0.45] node[midway, above, sloped, pos=0.5, yshift=-9.5pt] {\scalebox{0.6}{$w_{03|2}$}} (5);
        \draw[-] (4) to [out=0, in=20, looseness=2.0] node[midway, above, sloped, pos=0.5, yshift=-3.25pt] {\scalebox{0.6}{$w_{04|2}$}} (3);
        \draw[-] (4) to [out=255, in=-110, looseness=1.05] node[midway, above, sloped, pos=0.55, yshift=-9.75pt] {\scalebox{0.6}{$w_{02|2}$}} (6);
        \draw[-] (2) to [out=215, in=-35, looseness=0.45] node[midway, above, sloped, pos=0.5, yshift=-9.5pt] {\scalebox{0.6}{$w_{12|2}$}} (6);
        \draw[-] (2) to [in=225, out=135, looseness=0.45] node[midway, above, sloped, pos=0.5, yshift=-3.25pt] {\scalebox{0.6}{$w_{41|2}$}} (3);
        \draw[-] (5) to [out=105, in=357, looseness=0.60] node[midway, above, sloped, pos=0.5, yshift=-3.55pt] {\scalebox{0.6}{$w_{34|2}$}} (3);
        \draw[-] (3) to [out=185, in=75, looseness=0.50] node[midway, above, sloped, pos=0.5, yshift=-3.25pt] {\scalebox{0.6}{$w_{42|2}$}} (6);
        \draw[-] (6) to [out=275, in=280, looseness=1.20] node[midway, above, sloped, pos=0.90, yshift=-4.75pt] {\scalebox{0.6}{$w_{23|2}$}} (5);
        
    \end{tikzpicture} 
    }
    \vspace{-3.40em}
    \caption{Instance of the graph $G$ assuming three target areas (round solid nodes), one blade to cover (round dashed nodes), and two~\acp{UAV}. Round double nodes are the depots. Notably, we assume that $w_{ij|d} = w_{ji|d}$.}
    \label{fig:undirectedMultigraph}
\end{figure}
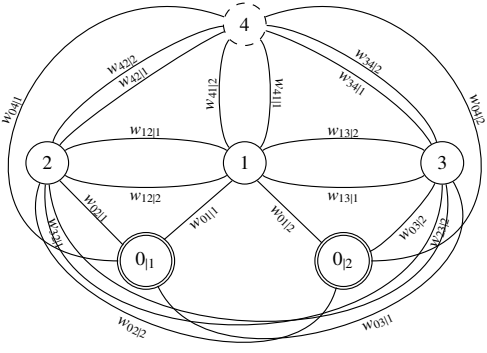

A directed weighted multigraph represented by the tuple $G = (\mathcal{V}, \mathcal{E}, \mathcal{W}, \mathcal{D})$ is used to formulate the~\ac{MILP}, as shown in Figure~\ref{fig:undirectedMultigraph}. The set of vertices $\mathcal{V}$ is defined as $\mathcal{T} \cup \mathcal{O}$, where $\mathcal{T}$  correspond to the target areas and the blade extreme points, and $\mathcal{O}=\{o_1, \dots, o_\delta\}$ comprises the depots where each~\ac{UAV} is located at $t_0$. The cardinalities of the sets are denoted by $\lvert \mathcal{T} \rvert = \tau$ and $\lvert \mathcal{O} \rvert = \delta$. The graph's edges and their weights are described by the sets $\mathcal{E}$ and $\mathcal{W}$, respectively. The set $\mathcal{D}$, as said before, contains $\delta$ available~\acp{UAV}. The graph connectivity is such that all vertices in $\mathcal{T}$ are fully connected to each other and to all vertices in the set $\mathcal{O}$, while the vertices in $\mathcal{O}$ are unconnected. Specifically, for any $\{i,j\} \in \mathcal{V}$ and $d \in \mathcal{D}$, the edge $e_{ij | d} \in \mathcal{E}$ connects vertices $i$ and $j$ using~\ac{UAV} $d$, while $w_{ij | d} \in \mathcal{W}$ denotes the corresponding weight. Notably, we assume that $w_{ij | d} = w_{ji | d}$ models the edge weights based on the~\acp{UAV}' time of flight given their dynamic constraints, which are heterogeneous and denoted by ${^d}\underline{v}^{(j)}$, ${^d}\underline{a}^{(j)}$, ${^d}\bar{v}^{(j)}$ and ${^d}\bar{a}^{(j)}$. 

The~\ac{UAV} motion primitives ${^d}\mathbf{S}^{(j)}$ (see Section~\ref{sec:STLMotionPlanner}) are employed to calculate the flight time for each~\ac{UAV} as it travels between target areas and blade extreme points, taking into account the physical limitations of the vehicles in terms of velocity (${^d}\underline{v}^{(j)}$ and ${^d}\bar{v}^{(j)}$) and acceleration (${^d}\underline{a}^{(j)}$ and ${^d}\bar{a}^{(j)}$). For brevity, further details for computing these times can be found elsewhere in~\cite{Mueller2015TRO}. 
The objective of the mission is to minimize the overall time. 

To indicate the number of times an edge is selected in the~\ac{MILP} solution, we assign an integer variable $z_{ij | d}$, where $z_{ij | d} \in \mathbb{N}$ for $i \in \mathcal{O}$ and $j \in \mathcal{T}$, with $d \in \mathcal{D}$, to each edge $e_{ij | d} \in \mathcal{E}$. 
We define an \textit{in-neighborhood}, denoted by $\mathcal{N}_i^\mathrm{in}$ the set of nodes having an edge entering $i$, that is $\mathcal{N}_i^\mathrm{in}=\{j\in \mathcal{V}: (j,i)\in\mathcal{E}\}$.
Similarly, we define an \textit{out-neighborhood} as the set of nodes having an entering edge which starts from $i$, that is
$\mathcal{N}_i^\mathrm{out}=\{j\in \mathcal{V}: (i,j)\in\mathcal{E}\}$.
Using these variables, we propose to formulate the~\ac{MILP} planning problem as follows:
\begin{subequations}\label{eq:MILP}
    \begin{align}
        &\minimize_{z_{ij|d}}
        { \sum\limits_{ \{i,j \} \in \mathcal{V}, \, i \neq j, \, d \in \mathcal{D}} \hspace{-0.30cm} w_{ij|d} \, z_{ij|d} } \label{subeq:objectiveFunction} \\
        %
        &\quad \text{s.t.} \hspace{0cm} \sum\limits_{i \in \mathcal{N}_j^\mathrm{in}} \hspace{-0.05cm} z_{ij|d} = \hspace{-0.25cm} \sum\limits_{i \in \mathcal{N}_j^\mathrm{out}} \hspace{-0.05cm} z_{ji|d}, \; \forall j \in \mathcal{T}, \; \forall d \in \mathcal{D}, \label{subeq:notAccumulating} \\ 
        &\qquad \, \; \sum\limits_{ d \in \mathcal{D}} \, \sum\limits_{ i \in \mathcal{N}_j^\mathrm{in}} \quad \hspace{-0.15cm} z_{ij|d} \geq 1, \; \forall j \in \mathcal{T}, \label{subeq:visitedOneUAV} \\ 
        &\qquad \hspace{-0.04cm} \sum\limits_{ j \in \mathcal{N}_{o_d}^\mathrm{out} } \hspace{-0.175cm} z_{o_d j|d} = 1, \; \forall d \in \mathcal{D}, \label{subeq:arrivalDepartureDepot} \\ 
        &\qquad \hspace{0.025cm} \sum\limits_{j \in \mathcal{N}_{o_d}^\mathrm{in} } \hspace{-0.175cm} z_{jo_d|d} = 1, \; \forall d \in \mathcal{D}. \label{subeq:arrivalDepartureSameDepot} 
     \end{align}
\end{subequations}

In the above formulae,~\eqref{subeq:objectiveFunction} is the objective function encoding the total flight time of the fleet of~\acp{UAV}. Notice that~\eqref{subeq:objectiveFunction} implicitly minimizes the fact that a drone passes by the same edges multiple (useless) times.~\eqref{subeq:notAccumulating} ensures that~\acp{UAV} do not accumulate in an area they reach.~\eqref{subeq:visitedOneUAV} ensures that all target areas and blade extreme points are visited at least once.~\eqref{subeq:arrivalDepartureDepot} and~\eqref{subeq:arrivalDepartureSameDepot} enforce the depot of each drone as its departure and arrival point. 
If disconnected subtours are provided in the~\ac{MILP} solution, we adopt the suboptimal decision of connecting them afterwards. Indeed,  since the~\ac{MILP} solution is only used to seed the final~\ac{STL} optimizer, this choice is not a significant concern and allows to save computational time~\ac{wrt} enforcing subtours elimination constraints. 
Notice that~\eqref{eq:MILP} can be adapted to meet the mission requirements, including additional futures, such as capacity restrictions for delivery missions, and removing tight limitations, such as the need to start and end at the depot position. Additionally, heuristics and metaheuristics techniques~\cite{Tan2021Access, Hillier2004} can be used to cope with the increasing number of constraints and variables. 

After obtaining the optimal assignment from the~\ac{MILP} planning problem~\eqref{eq:MILP}, the motion primitives described in~\cite{SilanoRAL2021} are used to generate dynamically feasible trajectories for each~\ac{UAV} in accordance with its optimal navigation sequence used to seed the~\ac{STL} optimizer.



\subsection{Event-based replanner}
\label{sec:eventBasedReplanner}

Environmental uncertainties and disturbances can cause mismatches between planned and actual drone trajectories (e.g., wind gusts or technical fault). In such cases, it is beneficial for the planner to compute a new trajectory that minimizes deviation from the optimal offline solution. To address this challenge, we present an improved event-based replanner~\cite{SilanoRAL2021} that continuously adjusts the drone trajectory to make up for lost time until the original time requirements are met. The revised trajectories have less strict vehicle limitations (${^d}\hat{v}^{(j)}$, ${^d}\hat{a}^{(j)}$, ${^d}\check{v}^{(j)}$, and ${^d}\check{a}^{(j)}$) than the original plan (${^d}\underline{v}^{(j)}$, ${^d}\underline{a}^{(j)}$, ${^d}\bar{v}^{(j)}$, and ${^d}\bar{a}^{(j)}$), assuming that the original plan was scheduled with ``conservative'' constraints to avoid stresses on the actuators. 

Let us introduce the revised velocity and acceleration constraints for each drone, characterized by their lower and upper limits, represented as ${^d}\hat{v}^{(j)}$, ${^d}\hat{a}^{(j)}$, ${^d}\check{v}^{(j)}$, and ${^d}\check{a}^{(j)}$. Here, ${^d}\hat{v}^{(j)} < {^d}\underline{v}^{(j)}$, ${^d}\hat{a}^{(j)} < {^d}\underline{a}^{(j)}$, ${^d}\check{v}^{(j)} > {^d}\bar{v}^{(j)}$, and ${^d}\check{a}^{(j)} > {^d}\bar{a}^{(j)}$. Let us also define the discrete time instants when events occur as $\bar{\mathbf{t}}=\mathrm{(\bar{t}_0, \dots, \bar{t}_E)}^\top \in \mathbb{R}^{E+1}$ and $\bar{t}_k$ as a generic entry of $\bar{\mathbf{t}}$. 
Finally, let ${^d}\tilde{\mathbf{p}}_k$ and ${^d}{\mathbf{p}}_k^\star$ represent the runtime and optimal positions of the $d$-th drone, respectively. Hence, we can define the event trigger condition for the planner as $\lVert {^d}\tilde{\mathbf{p}}_k - {^d}\mathbf{p}^\star_k \rVert > \eta$, where $\eta \in \mathbb{R}_{>0}$ is a threshold value set to achieve the best system behavior. Whenever this condition is met, a trigger is generated and the planner computes a new plan online using the motion planner in Sections~\ref{sec:specificationMapping} and~\ref{sec:milpEncoding} over the time interval $[\bar{t}_k, \hat{t}_k]$. Here, $\hat{t}_k$ denotes the time instance associated with the next task assignment before the trigger was generated. Note that only the affected drone has its trajectory recomputed, while the other functioning drones follow their original plans.

It is noteworthy that the computational effort needed for replanning is considerably less compared to initial planning, as it deals with only one \ac{UAV} and a reduced set of task regions (those yet to be visited). While exploring alternative strategies for replanning the routes of delayed drones may potentially yield better solutions in terms of robustness scores, it is crucial to highlight that our focus in this paper is on ensuring operational continuity, minimizing disruptions, and upholding safety in hazardous scenarios. 



\subsection{Attrition-aware planner}
\label{sec:attrititionAwarePlanner}

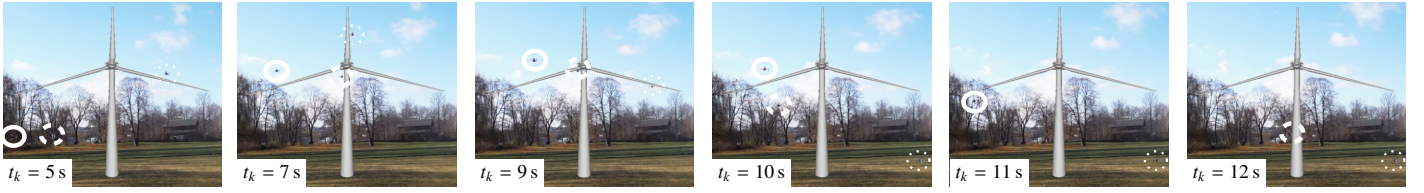
\begin{figure*}
    \centering
    \hspace*{-1.60cm}
\begin{subfigure}{0.155\columnwidth}
  \centering
  \scalebox{1.05}{
  \begin{tikzpicture}
    \node[anchor=south west,inner sep=0] (img) at (0,0) { 
    \adjincludegraphics[trim={{.325\width} {.08\height} {0.325\width} 
    {.39\height}},clip,scale=0.185]{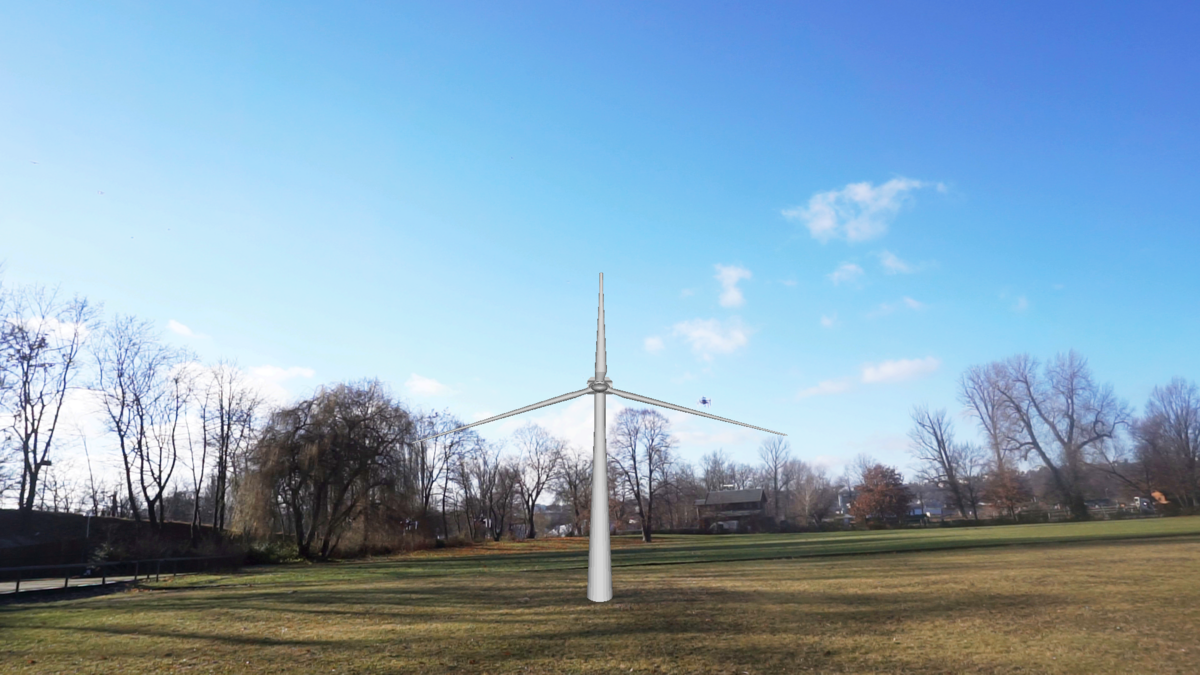}};
    \begin{scope}[x={(img.south east)},y={(img.north west)}]
      \draw [white, solid, ultra thick] (0.05, 0.275) circle (0.05); 
      \draw [white, dashed, ultra thick] (0.2250, 0.285) circle (0.05); 
      \draw [white, dotted, thick] (0.75, 0.61) circle (0.05); 
      \node[imglabel,text=black] (label) at (img.south west) {\scriptsize 
      $t_k=\SI{5}{\second}$};
    \end{scope}
  \end{tikzpicture}
  }
\end{subfigure}
\hspace*{1.575cm}
\begin{subfigure}{0.155\columnwidth}
  \centering
  \scalebox{1.05}{
  \begin{tikzpicture}
    \node[anchor=south west,inner sep=0] (img) at (0,0) { 
    \adjincludegraphics[trim={{.325\width} {.08\height} {0.325\width} 
    {.39\height}},clip,scale=0.185]{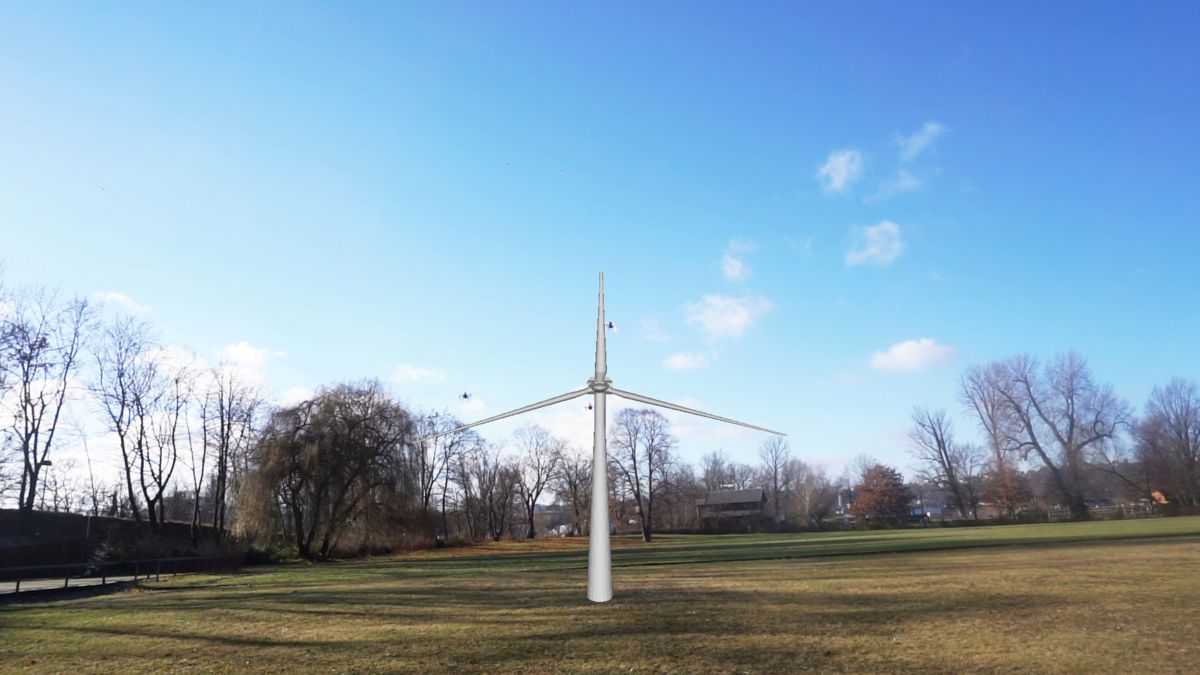}};
    \begin{scope}[x={(img.south east)},y={(img.north west)}]
      \draw [white, solid, ultra thick] (0.18, 0.625) circle (0.05); 
      \draw [white, dashed, ultra thick] (0.47, 0.595) circle (0.05); 
      \draw [white, dotted, thick] (0.525, 0.82) circle (0.05); 
      \node[imglabel,text=black] (label) at (img.south west) {\scriptsize 
      $t_k=\SI{7}{\second}$};
    \end{scope}
  \end{tikzpicture}
  }
\end{subfigure}
\hspace*{1.575cm}
\begin{subfigure}{0.155\columnwidth}
  \centering
  \scalebox{1.05}{
  \begin{tikzpicture}
    \node[anchor=south west,inner sep=0] (img) at (0,0) { 
    \adjincludegraphics[trim={{.325\width} {.08\height} {0.325\width} 
    {.39\height}},clip,scale=0.185]{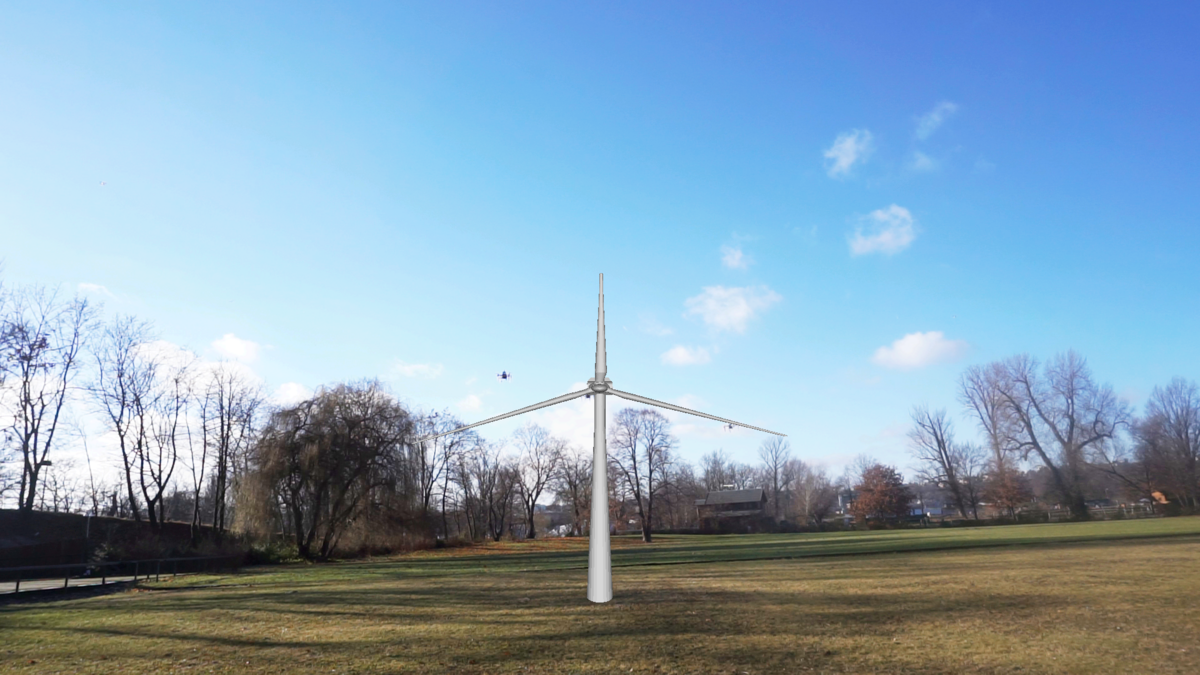}};
    \begin{scope}[x={(img.south east)},y={(img.north west)}]
      \draw [white, solid, ultra thick] (0.275, 0.685) circle (0.05); 
      \draw [white, dashed, ultra thick] (0.47, 0.64) circle (0.05); 
      \draw [white, dotted, thick] (0.81, 0.55) circle (0.05); 
      \node[imglabel,text=black] (label) at (img.south west) {\scriptsize 
      $t_k=\SI{9}{\second}$};
    \end{scope}
  \end{tikzpicture}
  }
\end{subfigure}
\hspace*{1.575cm}
\begin{subfigure}{0.155\columnwidth}
  \centering
  \scalebox{1.05}{
  \begin{tikzpicture}
    \node[anchor=south west,inner sep=0] (img) at (0,0) { 
    \adjincludegraphics[trim={{.325\width} {.08\height} {0.325\width} 
    {.39\height}},clip,scale=0.185]{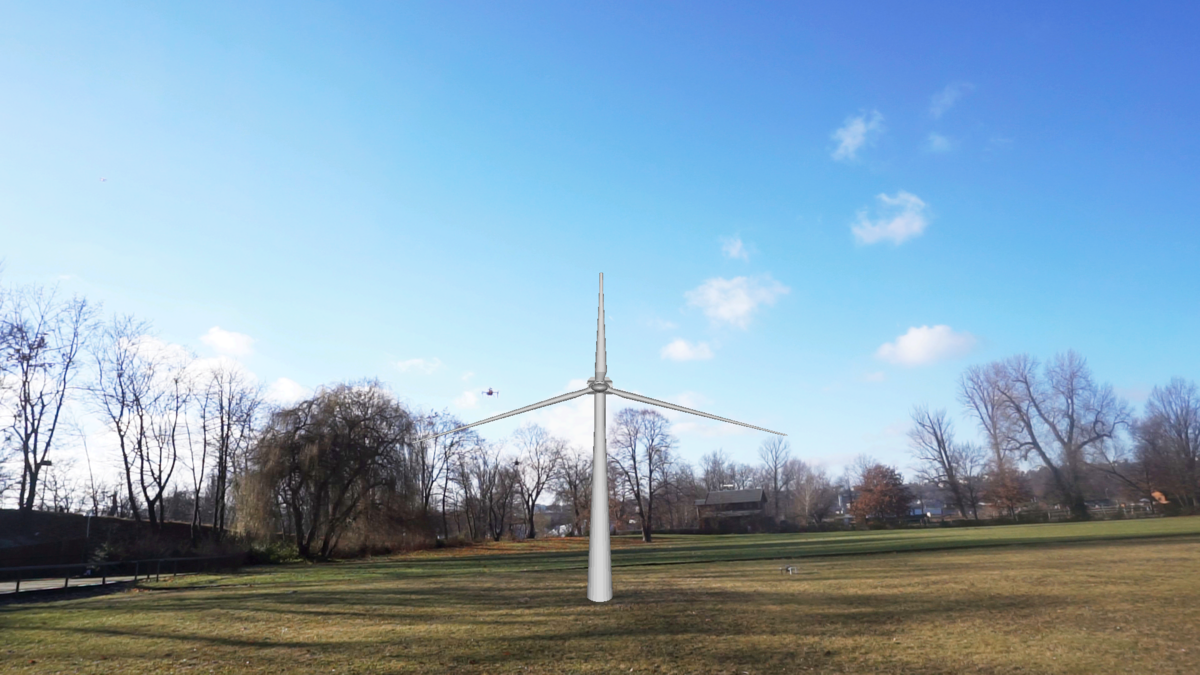}};
    \begin{scope}[x={(img.south east)},y={(img.north west)}]
      \draw [white, solid, ultra thick] (0.24, 0.64) circle (0.05); 
      \draw [white, dashed, ultra thick] (0.305, 0.445) circle (0.05); 
      \draw [white, dotted, thick] (0.94, 0.145) circle (0.05); 
      \node[imglabel,text=black] (label) at (img.south west) {\scriptsize 
      $t_k=\SI{10}{\second}$};
    \end{scope}
  \end{tikzpicture}
  }
\end{subfigure}
\hspace*{1.575cm}
\begin{subfigure}{0.155\columnwidth}
  \centering
  \scalebox{1.05}{
  \begin{tikzpicture}
    \node[anchor=south west,inner sep=0] (img) at (0,0) { 
    \adjincludegraphics[trim={{.325\width} {.08\height} {0.325\width} 
    {.39\height}},clip,scale=0.185]{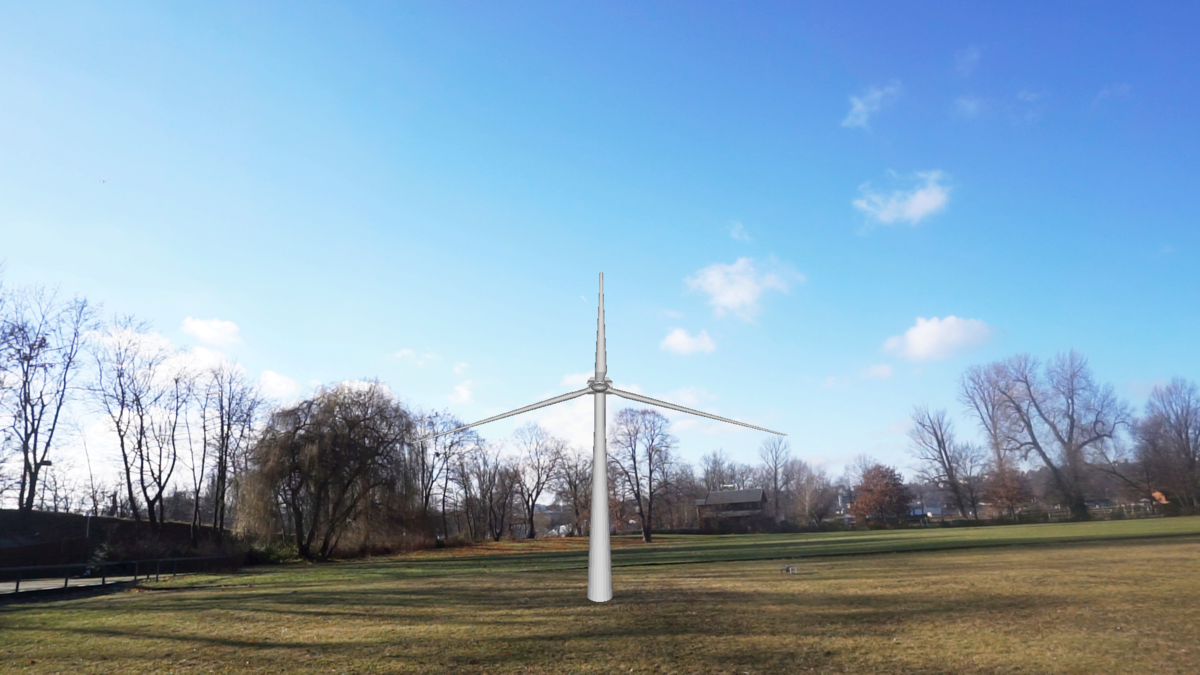}};
    \begin{scope}[x={(img.south east)},y={(img.north west)}]
      \draw [white, solid, ultra thick] (0.115, 0.46) circle (0.05); 
      \draw [white, dotted, thick] (0.94, 0.145) circle (0.05); 
      \node [imglabel,text=black] (label) at (img.south west) {\scriptsize 
      $t_k=\SI{11}{\second}$};
    \end{scope}
  \end{tikzpicture}
  }
\end{subfigure}
\hspace*{1.575cm}
\begin{subfigure}{0.155\columnwidth}
  \centering
  \scalebox{1.05}{
  \begin{tikzpicture}
    \node[anchor=south west,inner sep=0] (img) at (0,0) { 
    \adjincludegraphics[trim={{.325\width} {.08\height} {0.325\width} 
    {.39\height}},clip,scale=0.185]{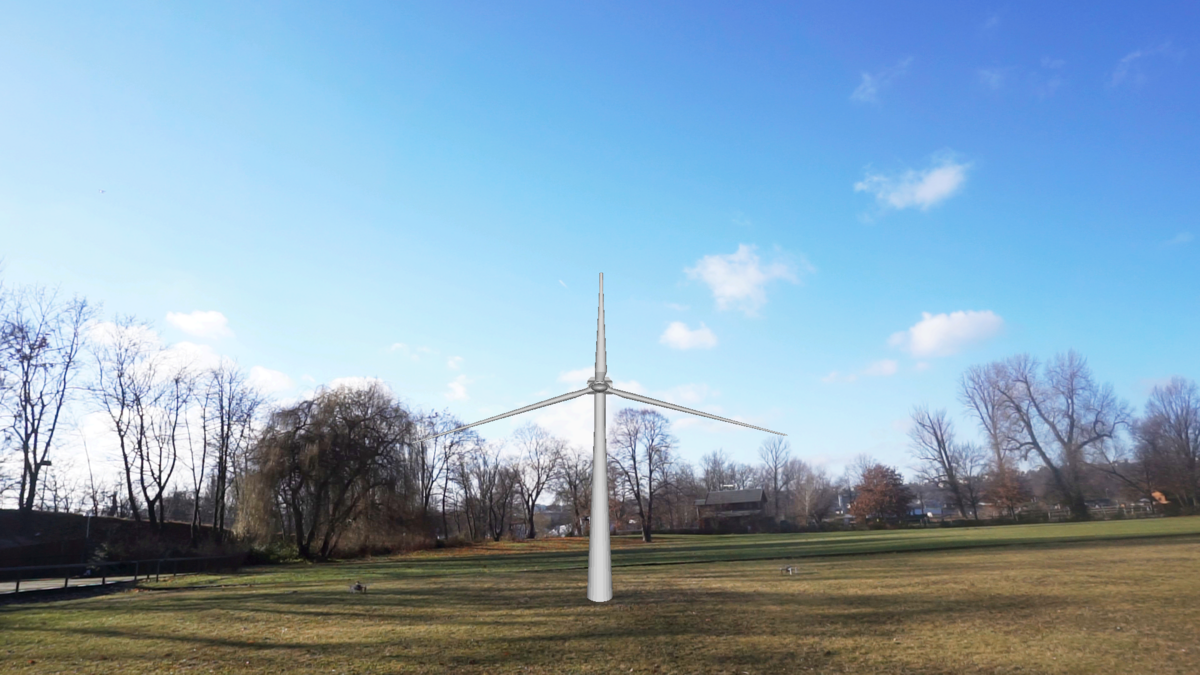}};
    \begin{scope}[x={(img.south east)},y={(img.north west)}]
      \draw [white, dashed, ultra thick] (0.48, 0.295) circle (0.05); 
      \draw [white, dotted, thick] (0.94, 0.145) circle (0.05); 
      \node [imglabel,text=black] (label) at (img.south west) {\scriptsize 
      $t_k=\SI{12}{\second}$};
    \end{scope}
  \end{tikzpicture}
  }
\end{subfigure}
    \vspace{-1.0mm}
    \caption{Experiment snapshots. Solid, dashed, and dotted circles indicate ``drone1", ``drone2", and ``drone3", respectively. The wind turbine was virtually projected within the~\acp{UAV}' camera frames to convey its proximity to the infrastructure.}
    \label{fig:experimentsTimeLine}
\end{figure*}

In collaborative missions, conflicting tasks can lead to an infeasible problem for the actual~\ac{STL} robustness score. For example, a drone may be tasked with surveying areas that lack sufficient safety margins for the entire mission duration. To tackle this issue, incorporating user preferences into the planner formulation can be beneficial. The goal is to generate a feasible vehicle trajectory that prioritizes mission specifications, while ensuring their temporal satisfaction. A generalized robustness score approach for~\ac{STL} robust semantics~\cite{MehdipourLCSS2021, Cardona2023ACM, CardonaECC2023} can be employed to formally capture requirements using weights. Rather than just focusing on boundary conditions, this allows for the assignment of time priorities when the specification is satisfied.

This addresses some limitations of the traditional robustness score, such as the inability to specify when a proposition should be preferentially satisfied, such as at the beginning or just before the end of the optimization. As an example, we consider the generalized robustness score for the logical operator \textit{and} ($\wedge$) (see Section~\ref{sec:weightedSignalTemporalLogic}):
\begin{equation*}\label{eq:examplewSTL} 
   {^\omega}\rho_{\varphi_1 \wedge \varphi_2}(\bm{\omega}, \mathbf{x}, t_k) = \min \left(\omega_1\rho_{\varphi_1} (\mathbf{x}, t_k), \omega_2\rho_{\varphi_2} (\mathbf{x}, t_k) \right) ,
\end{equation*}
where ${^\omega}\rho_{\varphi_1 \wedge \varphi_2}(\bm{\omega}, \mathbf{x}, t_k)$ refers to the weighted version of the~\ac{STL} robustness formula $\rho_{\varphi_1 \wedge \varphi_2}(\mathbf{x}, t_k)$.  The variable $\bm{\omega} = (\omega_1, \omega_2)^\top$, with $\{\omega_1,\omega_2\} \in \mathbb{R}_{>0}$, represents the vector of weights applied to each predicate of the~\ac{STL} formula $\varphi = \varphi_1 \wedge \varphi_2$. 

Thus, we can express the weighted generalization of the standard smooth robustness of the \ac{STL} formula $\varphi = \varphi_1 \wedge \varphi_2$, denoted by ${^\omega}\tilde{\rho}_\varphi(\bm{\omega}, \mathbf{x}, t_k)$, as:
\begin{equation*}
   {^\omega}\tilde{\rho}_{\varphi_1 \wedge \varphi_2}(\bm{\omega}, \mathbf{x}, t_k) = \min_{i = \{1, 2\} } \left \{ \left( \left( \frac{1}{2} - \bar{\omega}_i \right) \mathrm{sign}(\tilde{\rho}_{\varphi_i}) + \frac{1}{2} \right) \tilde{\rho}_{\varphi_i} \right \},
\end{equation*}
where $\bar{\omega}_i  = \omega_i / ( \mathbf{I}_2 \bm{\omega} )$ are normalized weights, with $\mathbf{I}_2 \in \mathbb{R}^{2 \times 2}$ being the identity matrix. The complete syntax and semantics of the generalized robustness score approach can be found in~\cite{MehdipourLCSS2021, Cardona2023ACM, CardonaECC2023}. Due to space limitations, full details of the approach are not reported here. 

Hence, we propose to formulate problem~\eqref{eq:optimizationProblemMotionPrimitives} by replacing the smooth approximation of the standard~\ac{STL} formula $\tilde{\rho}_\varphi(\mathbf{x}, t_k)$ with its weighted version ${^\omega}{\tilde{\rho}}_\varphi (\bm{\omega},\mathbf{x}, t_k)$ as follows:
\begin{equation}\label{eq:optimizationProblemWeighted}
    \begin{split}
    &\maximize_{{^d}\mathbf{p}^{(j)}, {^d}\mathbf{v}^{(j)},\,{^d}\mathbf{a}^{(j)}\atop d \in \mathcal{D}} \;\;
    {^\omega}{\tilde{\rho}}_\varphi(\bm{\omega}, \mathbf{p}, \mathbf{v} ) \\
    &\qquad \,\;\, \text{s.t.}~\quad\, {^d}\underline{v}^{(j)} \leq {^d}v^{(j)}_k \leq {^d}\bar{v}^{(j)}, \\
    &\,\;\;\;\, \qquad \qquad {^d}\underline{a}^{(j)} \leq {^d}a^{(j)}_k \leq {^d}\bar{a}^{(j)}, \\
    &\,\;\;\;\, \qquad \qquad {^\omega}\tilde{\rho}_\varphi (\bm{\omega}, {^d}\mathbf{p}^{(j)}, \, {^d}\mathbf{v}^{(j)}) \geq \zeta, \\
    &\,\;\;\;\, \qquad \qquad {^d}\mathbf{S}^{(j)}, \forall k=\{0,1, \dots, N-1\}
    \end{split}.
\end{equation}

In this way, we can ensure the satisfaction of collaborative missions while considering prioritized specifications to avoid issues that may arise from conflicting tasks during trajectory planning. For instance, if the original problem~\eqref{eq:optimizationProblemMotionPrimitives} is unable to find a solution due to conflicting tasks, adjusting the weights $\bm{\omega}$ of the specifications, such as prioritizing safety (${^d}\varphi_\mathrm{ws}$, ${^d}\varphi_\mathrm{obs}$, ${^d}\varphi_\mathrm{dis}$) and mission completion requirements (${^d}\varphi_\mathrm{hm}$) over task requirements (${^d}\varphi_\mathrm{tr}$ and ${^d}\varphi_\mathrm{bla}$), could lead to safer trajectories. However, this comes at the expense of a reduction in the overall robustness score $\rho_\varphi(\mathbf{x}, t_k)$.



\section{Experimental Results}
\label{sec:experimentalResults}

The proposed planning approach's validity and effectiveness were assessed through MATLAB and Gazebo simulations, along with field experiments in a mock-up scenario. Initially, numerical simulations were performed in MATLAB without explicitly modeling actual vehicle dynamics and trajectory tracking controllers. This approach allowed us to evaluate the planning algorithm's performance and gain valuable insights into its behavior. Subsequently, to further validate the generated trajectories and leverage the benefits of software-in-the-loop simulations~\cite{Baca2020mrs, Silano2019SMC}, we conducted additional simulations using the Gazebo robotics simulator. Ultimately, field experiments conducted in a mock-up scenario closely resembling real-world conditions demonstrated the practical applicability of the proposed method.

The experiments aimed to demonstrate several key aspects: (i) the alignment of planned trajectories with mission requirements, (ii) the necessity for the \ac{STL} motion planner to fulfill mission specifications, and to show where the \ac{MILP} alone is insufficient, (iii) the ability to replan missions in response to unexpected disturbances, (iv) a comparison of solutions with and without the attrition-aware planner, and (v) the feasibility of the method in real-world scenarios.

MATLAB was used to code the optimization method, with the~\ac{MILP} formulated using the CVX framework\footnote{\url{http://cvxr.com/cvx/}} and the~\ac{STL} motion planner using the CasADi library\footnote{\url{https://web.casadi.org/}} and IPOPT\footnote{\url{https://coin-or.github.io/Ipopt/}} as a solver. The simulations were performed on a computer running Ubuntu 20.04 with an i7-8565U processor (\SI{1.80}{\giga\hertz}) and $32$GB of RAM.  Videos with the experiments and the numerical simulations in MATLAB and Gazebo are available at~\url{http://mrs.felk.cvut.cz/milp-stl}. Figure~\ref{fig:experimentsTimeLine} shows experiment snapshots with virtualized representations of the wind turbine.



\subsection{Wind turbine inspection}
\label{sec:windTurbineInspection}

\begin{figure}[tb]
    \begin{center}
    \vspace*{-2em}
    \hspace*{10mm}
        \adjincludegraphics[trim={{.21\width} {.05\height} {0.0\width}	{.0\height}},clip,scale=0.25]{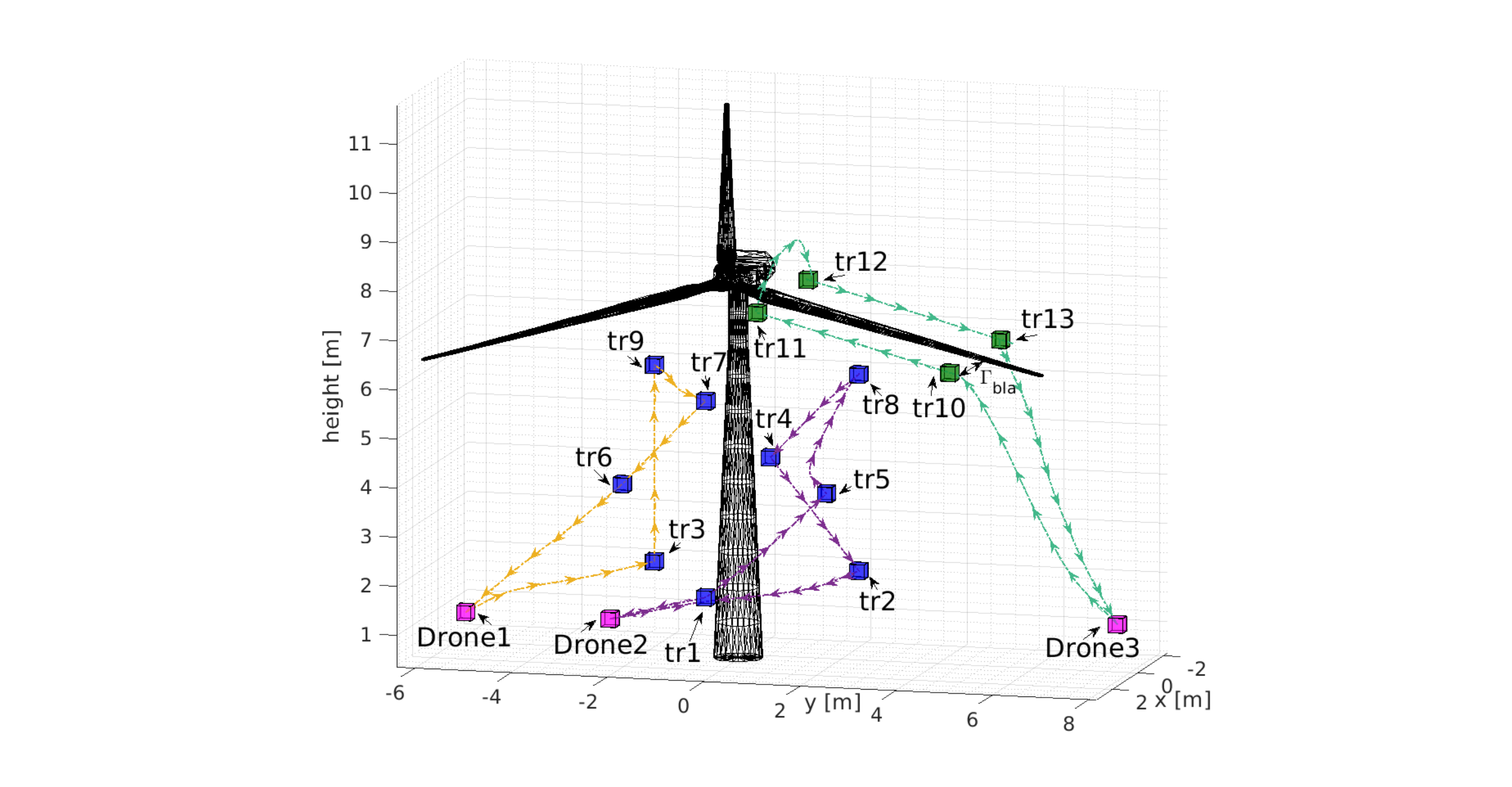}
	\vspace*{-10mm}
    \end{center}
    \caption{Wind turbine inspection scenario along with planner trajectories. Arrows depict the path followed by the drones throughout the mission.}
    \label{fig:inspectionScenario}
\end{figure}

We evaluated our planning approach using the wind turbine inspection scenario described in Section~\ref{sec:problemDescription} and three drones. The scenario included nine target areas for the pylon inspection task and a single blade for the blade inspection task, as shown in Figure~\ref{fig:inspectionScenario}, in a mock-up area of \SI{8}{\meter} $\times$ \SI{20}{\meter} $\times$ \SI{14}{\meter}. We chose to focus our inspection efforts on a single blade, as it provided sufficient scope to thoroughly test the effectiveness and feasibility of our methodology. The optimization problem was run with the parameter values listed in Table~\ref{tab:tableParamters}. To ensure a reasonable computation time, short symbolic mission time intervals ($T_\mathrm{ins}$, $T_\mathrm{bla}$, and $T_N$) were used. During the inspection operation, the heading angles of the drones (${^d}\psi$) were adjusted and aligned with the displacement direction when moving towards the target areas (pylon inspection) or the blade (blade inspection). Furthermore, during blade inspection (${^d}\varphi_\mathrm{bla}$), drones maintained their heading toward the blade surface between its two extremes, the leading edge and rotor shaft. Additionally, during pylon inspection (${^d}\varphi_\mathrm{tr}$), the drone's heading remained constant at a specific value for the inspection duration ($T_\mathrm{ins}$), chosen per each target based on the particular pylon area under surveillance.

\begin{figure}[tb]
    \centering
    \input{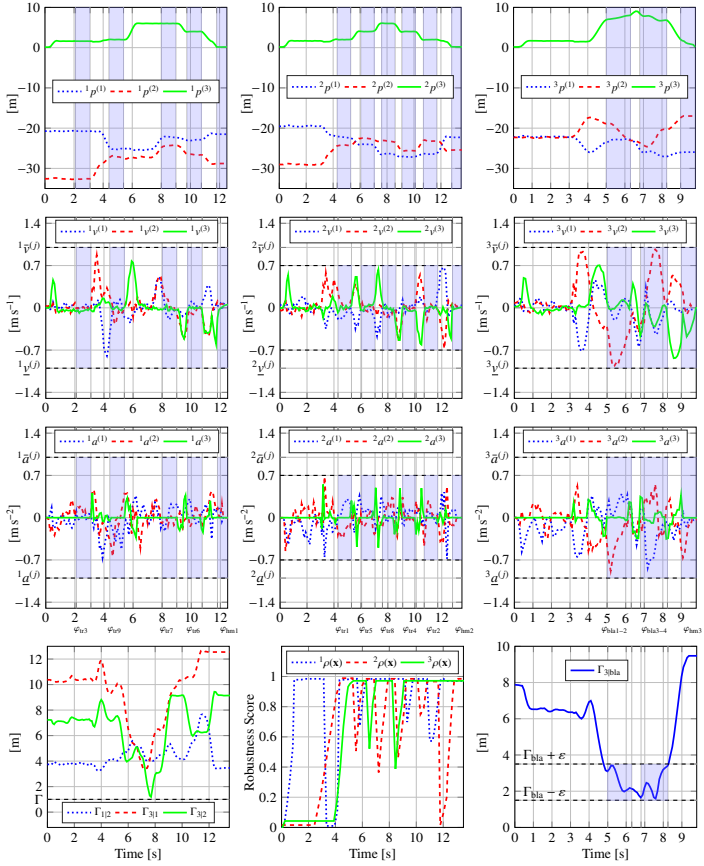}
    \vspace{-1.85mm}
    \caption{Position, velocity, acceleration, mutual safety distance ($\Gamma_{n|m}$), and distance from the blade $\Gamma_\mathrm{3|bla}$ surface. Blue-colored time windows indicate the pylon and blade inspection operations.}
    \label{fig:plotExperimentalResults}
\end{figure}

The planned trajectories, along with the wind turbine, target areas, and blade extreme points, are depicted in Figure~\ref{fig:inspectionScenario}. The wind turbine is \SI{12}{\meter} in height and \SI{12}{\meter} in width, with a blade surface extension of \SI{7}{\meter} starting from the rotor shaft to the leading edge. The small size of the wind turbine served two main purposes in our experiments. Firstly, it ensured that our experiments remained self-contained, conducted within a mock-up scenario to demonstrate the validity and effectiveness of our proposed planning approach. Secondly, it is worth noting that the experiments were conducted in an area where drone flight over $\SI{20}{\meter}$ in height was not permitted.  The optimization problem required \SI{145}{\second} to solve and \SI{4}{\second} to find an initial guess solution. Real-world experiments confirm the alignment of planned trajectories with mission requirements, as evidenced by the results in Figure~\ref{fig:plotExperimentalResults}. The results indicate that the distance between vehicles remains above the specified threshold value $\Gamma_\mathrm{dis}$ at all times, and the velocity and acceleration of each vehicle remains within the allowed bounds of $[{^d}\underline{v}^{(j)}, {^d}\bar{v}^{(j)}]$ and $[{^d}\underline{a}^{(j)}, {^d}\bar{a}^{(j)}]$, respectively. Additionally, throughout the blade inspection, the vehicle always maintain a certain distance ($\Gamma_{3|\mathrm{bla}} \in (\Gamma_\mathrm{bla} - \varepsilon, \Gamma_\mathrm{bla} + \varepsilon)$). Note that, for simplicity, we assume symmetric velocity and acceleration bounds for each drone, i.e., $\lvert {^d}\underline{v}^{(j)} \rvert = \lvert {^d}\bar{v}^{(j)} \rvert$ and $\lvert {^d}\underline{a}^{(j)} \rvert = \lvert {^d}\bar{a}^{(j)} \rvert$.

\begin{table}[tb]
    \centering
    \begin{adjustbox}{max width=0.98\columnwidth}
        \begin{tabular}{c|c|c|c|c|c}
          \hline
          \textbf{Description} & \textbf{Sym.} & \textbf{Value} & \textbf{Description} & \textbf{Sym.} & \textbf{Value}
          \\
          \hline \hline
          Sampling period & $T_s$ & \SI{0.05}{[\second]} & Tunable parameter & $\lambda$ & $\SI{10}{[-]}$ \\
          Safety distance & $\Gamma_\mathrm{dis}$ & \SI{1}{[\meter]} & Blade distance & $\Gamma_\mathrm{bla}$ & \SI{2.5}{[\meter]} \\
          Max velocity & $\lVert {^d}\bar{v}^{(j)} \rVert$ & $\{1, 0.7, 1\}$\si{[\meter\per\second]} & Max accel. & $\lVert {^d}\bar{a}^{(j)} \rVert$ & $\{1, 0.7, 1\}$\si{[\meter\per\square\second]} \\ 
          %
          %
          Trigger. cond & $\eta$ & $\SI{1}{[\meter]}$  & Blade margin & $\varepsilon$ & $\SI{1}{[\meter]}$ \\
          Rev. max velocity & $\lVert {^d}\hat{v}^{(j)} \lVert$ & \SI{2}{[\meter\per\second]} & Rev. max accel. & $\lVert {^d}\check{a}^{(j)} \rVert$ & \SI{5}{[\meter\per\square\second]} \\
          Weight $\varphi_\mathrm{tr}$ & $\omega_{\varphi_\mathrm{tr}}$ & $\SI{1}{[-]}$ &  Weight $\varphi_\mathrm{bla}$ & $\omega_{\varphi_\mathrm{bla}}$ & $\SI{1}{[-]}$ \\
          Weight $\varphi_\mathrm{ws}$ & $\omega_{\varphi_\mathrm{ws}}$ & $\SI{2}{[-]}$ & Weight $\varphi_\mathrm{hm}$ & $\omega_{\varphi_\mathrm{hm}}$ & $\SI{1}{[-]}$ \\
          Weight $\varphi_\mathrm{obs}$ & $\omega_{\varphi_\mathrm{obs}}$ & $\SI{4}{[-]}$ &  Pylon insp. & $T_\mathrm{ins}$ & $\SI{1}{[\second]}$ \\
          Weight $\varphi_\mathrm{dis}$ & $\omega_{\varphi_\mathrm{dis}}$ & $\SI{3}{[-]}$ & Mission time & $T_N$ & $\SI{13}{[\second]}$ \\
          \ac{STL} safety margin & $\zeta$ & $\SI{0.2}{[-]}$ & Blade insp. & $T_\mathrm{bla}$ & $\SI{1.5}{[\second]}$ \\
          \hline
        \end{tabular}
    \end{adjustbox}
    \caption{Optimization problem parameter values.}
    \label{tab:tableParamters}
\end{table}



\subsection{Comparative analysis with the initial guess solution}
\label{sec:comparativeAnalysis}

\begin{figure}[tb]
    \centering
    \vspace*{-4mm}
    \hspace*{10mm}
		\adjincludegraphics[trim={{.21\width} {.05\height} {0.0\width} {.0\height}},clip,scale=0.25]{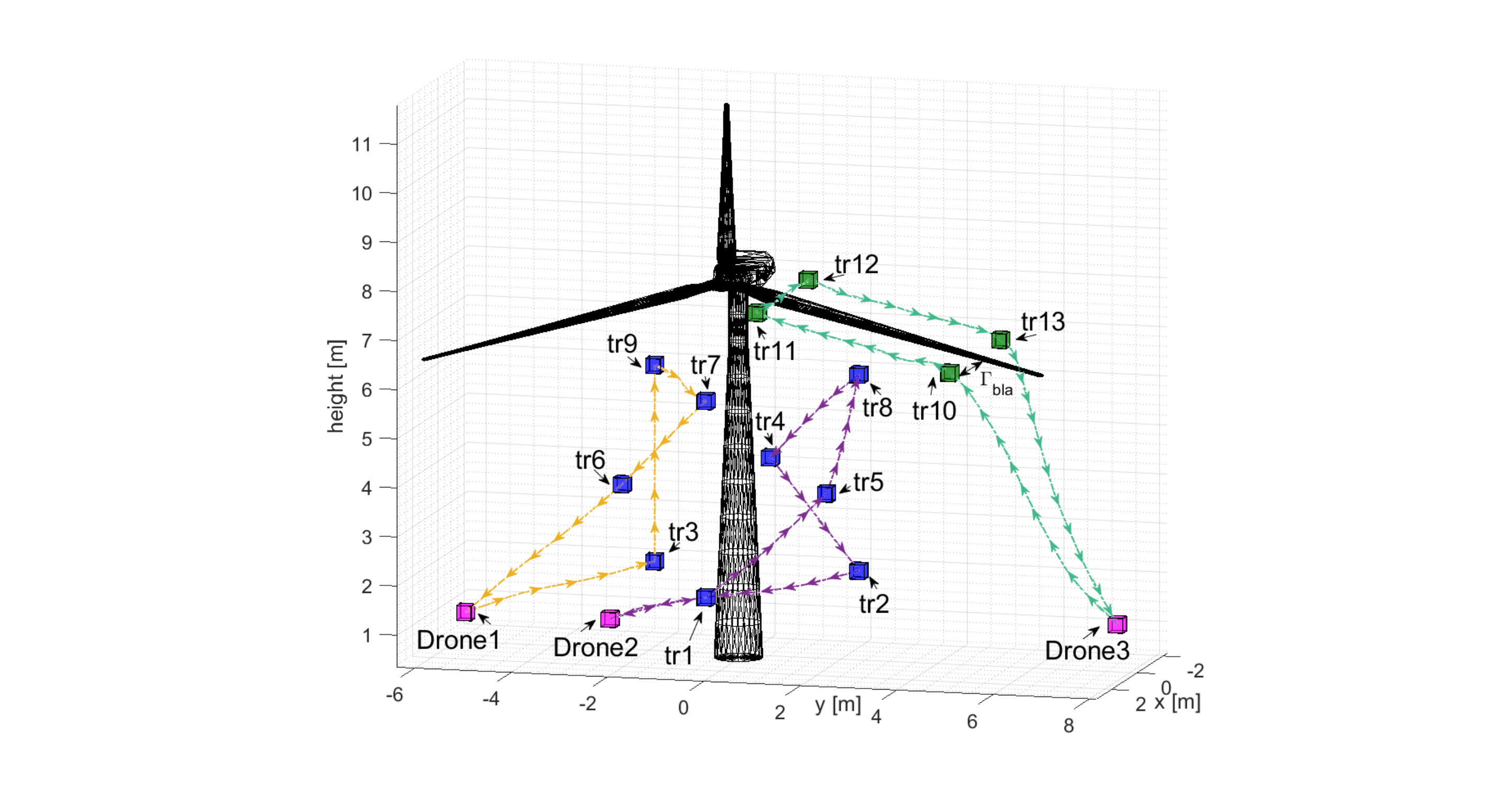}
    \vspace*{-8mm}
    \caption{Trajectories obtained solely through the~\ac{MILP} formulation, without using the~\ac{STL} planner.}
    \label{fig:windTurbineInspectionMILP}
\end{figure}

In this section, we assess the hierarchical planner's effectiveness in solving the collaborative inspection problem. Notably, the~\ac{STL} optimization problem~\eqref{eq:optimizationProblemMotionPrimitives} is nonlinear and non-convex, requiring an initial guess for convergence to a feasible solution. As a result, we compare the~\ac{MILP} solution~\eqref{eq:MILP} with the solution obtained from the~\ac{STL} optimization, seeded with the~\ac{MILP} solution since we cannot compare them directly. The solver was indeed not able to converge for the pure \ac{STL} problem with generic initial condition, so highlighting the importance of having an~\ac{MILP} problem solution for warm starting the~\ac{STL} solution computation. We evaluate compliance with the mission requirements and how the hierarchical planner addresses nonlinear complexities such as obstacle avoidance, safety distance, and time requirements, which the~\ac{MILP} alone cannot handle. Figure~\ref{fig:windTurbineInspectionMILP} depicts the trajectories generated using motion primitives~\cite{SilanoRAL2021} and the waypoint sequences assigned by the~\ac{MILP} to each~\ac{UAV}.

Upon closer inspection, it is somewhat evident that the~\ac{MILP} formulation neglects vehicle accelerations $\mathbf{a}^{(j)}$, which are used for controlling the vehicles' motion. This may yield impractical trajectories or contain sharp turns and corners, potentially deviating from mission requirements, posing safety hazards, and resulting in high energy consumption. Furthermore, the~\ac{MILP} formulation also fails to consider the distance constraints (${^d}\varphi_\mathrm{dis}$ and ${^d}\varphi_\mathrm{bla}$) due to the computational burden of accounting for vehicle dynamics. Thus, changing the $\Gamma_\mathrm{dis}$~\eqref{subeq:keepDistance} and $\Gamma_\mathrm{bla}$~\eqref{subeq:bladeInspection} values may require a completely different set of optimal sequences (output of the~\ac{MILP} solver) for the final solution of the problem. Additionally, the~\ac{MILP} formulation does not address obstacle avoidance, which can be observed in Figure~\ref{fig:windTurbineInspectionMILP} with the trajectory crossing the wind turbine. In contrast, Figure~\ref{fig:inspectionScenario} shows how the~\ac{STL} formulation fine-tunes the initial guess solution to meet the mutual safety distance constraint and other mission requirements.

\begin{figure}[tb]
    \centering
    \hspace*{-5mm}
	\begin{subfigure}{\columnwidth}
        \adjincludegraphics[trim={{.05\width} {.32\height} {0.0\width} {.335\height}},clip,scale=0.21]{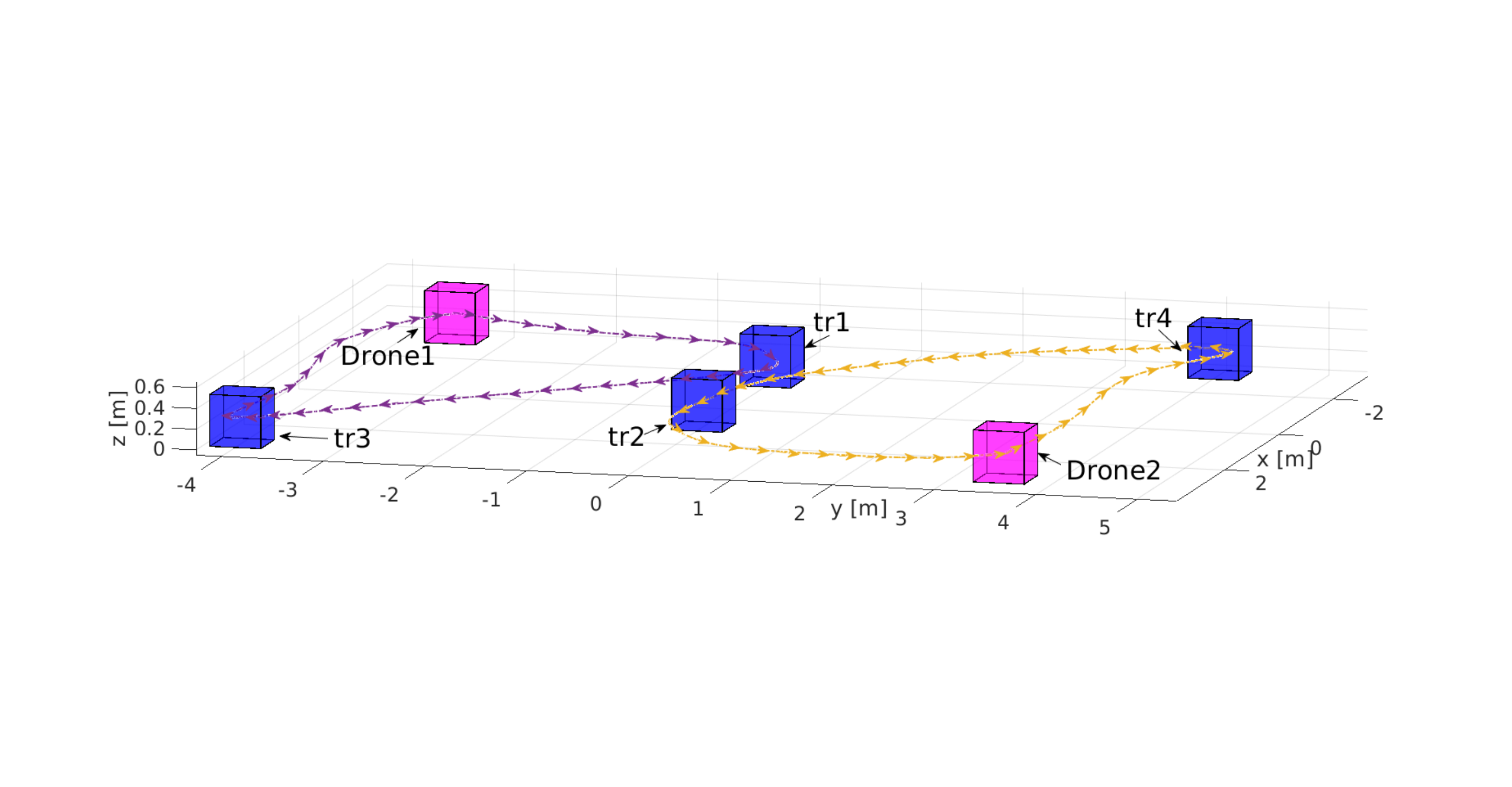}
        \vspace{-2.0em}
        \caption{}
        \label{subfig:MILPsubfig}
	  \end{subfigure}
	  \\
	  %
	  \begin{subfigure}{\columnwidth}
        \adjincludegraphics[trim={{.05\width} {.32\height} {0.0\width} {.335\height}},clip,scale=0.21]{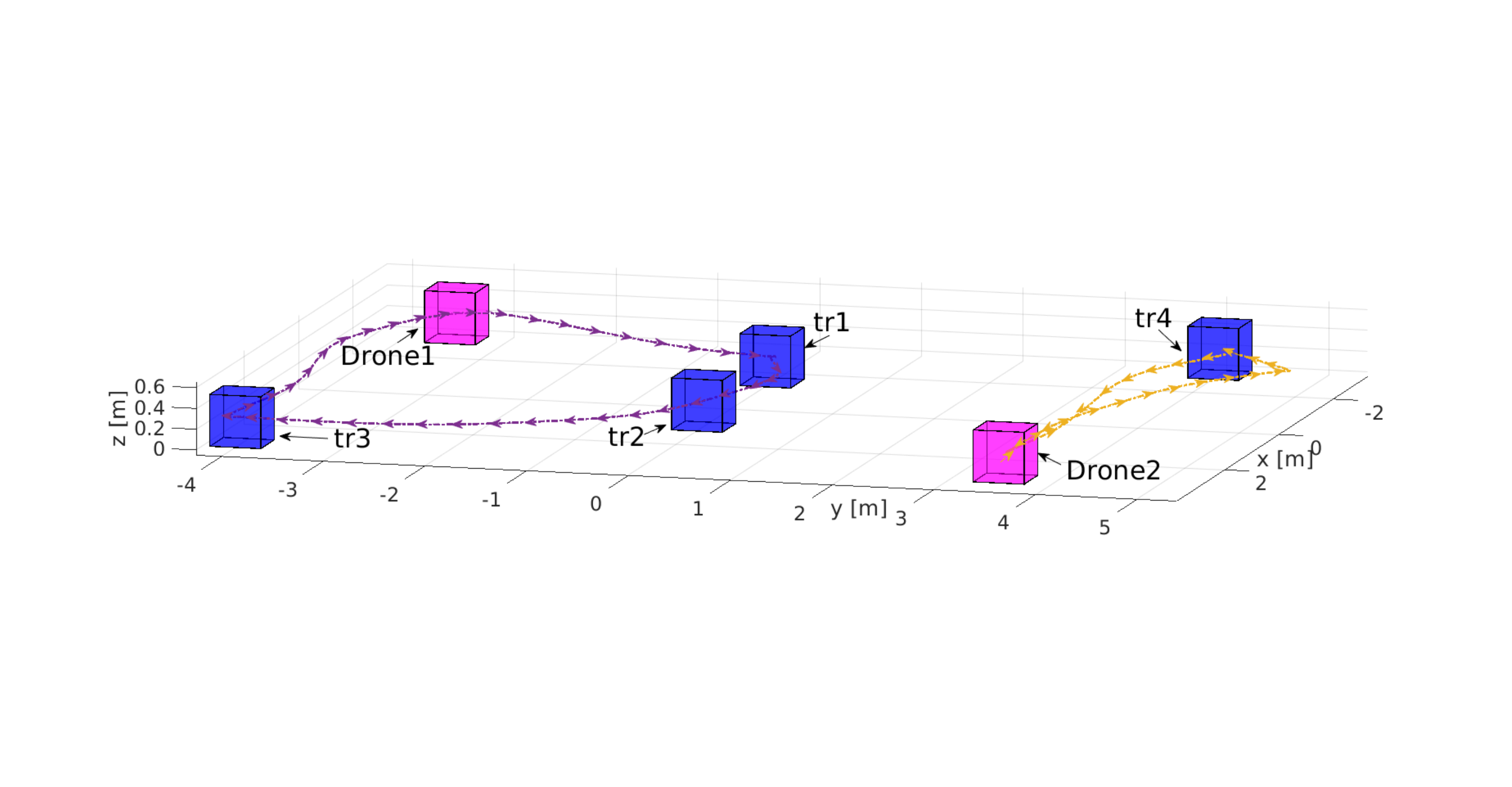}    	
        \vspace{-2.00em}
        \caption{}
        \label{subfig:STLsubfig}
        \end{subfigure}
        \vspace{-0.8em}
    \caption{Simple scenario showing the independence of the final~\ac{STL} solution~\eqref{subfig:STLsubfig} from the~\ac{MILP} initial guess~\eqref{subfig:MILPsubfig}.}
    \label{fig:dummyScenario}
\end{figure}

Finally, the differences between the~\ac{STL} and~\ac{MILP} solutions not only affect the shape of trajectories, but also the sequence of targets to visit. To illustrate this point, Figure~\ref{fig:dummyScenario} presents a simple scenario involving two~\acp{UAV} tasked with visiting a set of target areas within a specified time interval. Here, the~\ac{MILP} initial guess (Figure~\ref{subfig:MILPsubfig}) assigns the workload between the~\acp{UAV} minimizing the total flight time, but ignores the vehicle dynamic and time constraints. Conversely, the~\ac{STL} planner (Figure~\ref{subfig:STLsubfig}) reassigns targets to meet all mission specifications, showing the flexibility of the hierarchical planner. 

\subsection{Attrition-aware and event-triggered replanner}
\label{sec:attritionAwareEventTriggeredReplanner}

\begin{figure}[tb]
    \centering
    \vspace*{-3mm}
    \hspace*{10mm}
	\adjincludegraphics[trim={{.21\width} {.05\height} {0.0\width} {.0\height}},clip,scale=0.25]{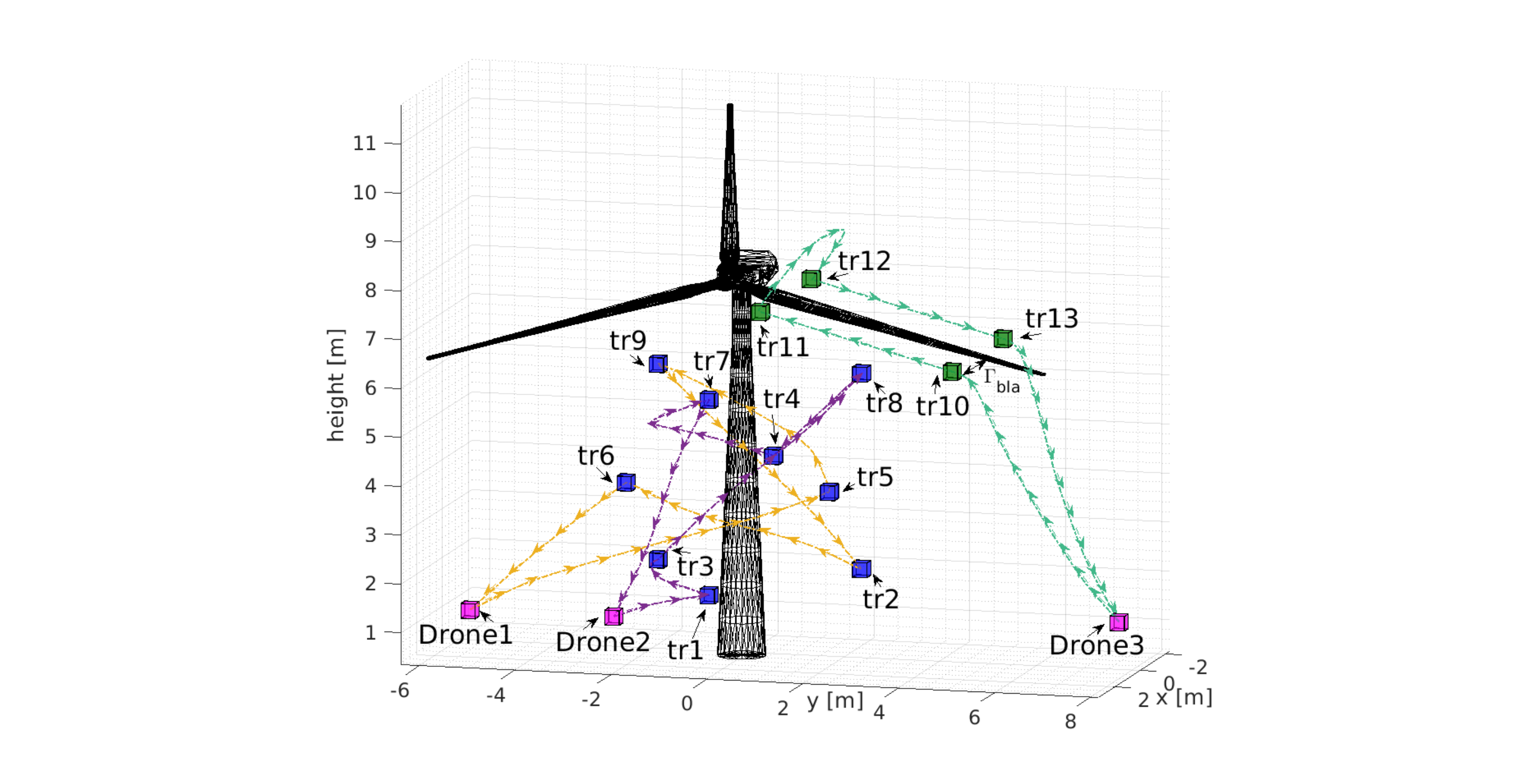}
	\vspace*{-8mm}
    \caption{Wind turbine inspection scenario considering the attrition-aware motion planner.}
    \label{fig:attritionAwarePlanner}
\end{figure}

\begin{figure}[tb]
    \centering
    \input{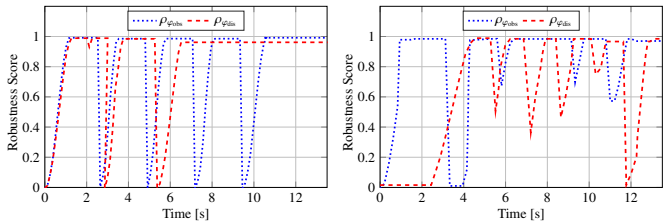}
    \vspace{-3.0mm}
    \caption{Robustness score profiles for $\varphi_\mathrm{obs}$ and $\varphi_\mathrm{dis}$ specifications. From left to right: data when considering the ``basic”~\eqref{eq:optimizationProblemMotionPrimitives} and the attrition-aware~\eqref{eq:optimizationProblemWeighted} motion planner, respectively.}
    \label{fig:distanceCostFunctionWSTL}
\end{figure}

\begin{figure}[tb]
    \centering
    \vspace*{-3mm}
    \hspace*{10mm}
	\adjincludegraphics[trim={{.21\width} {.05\height} {0.0\width} {.06\height}},clip,scale=0.25]{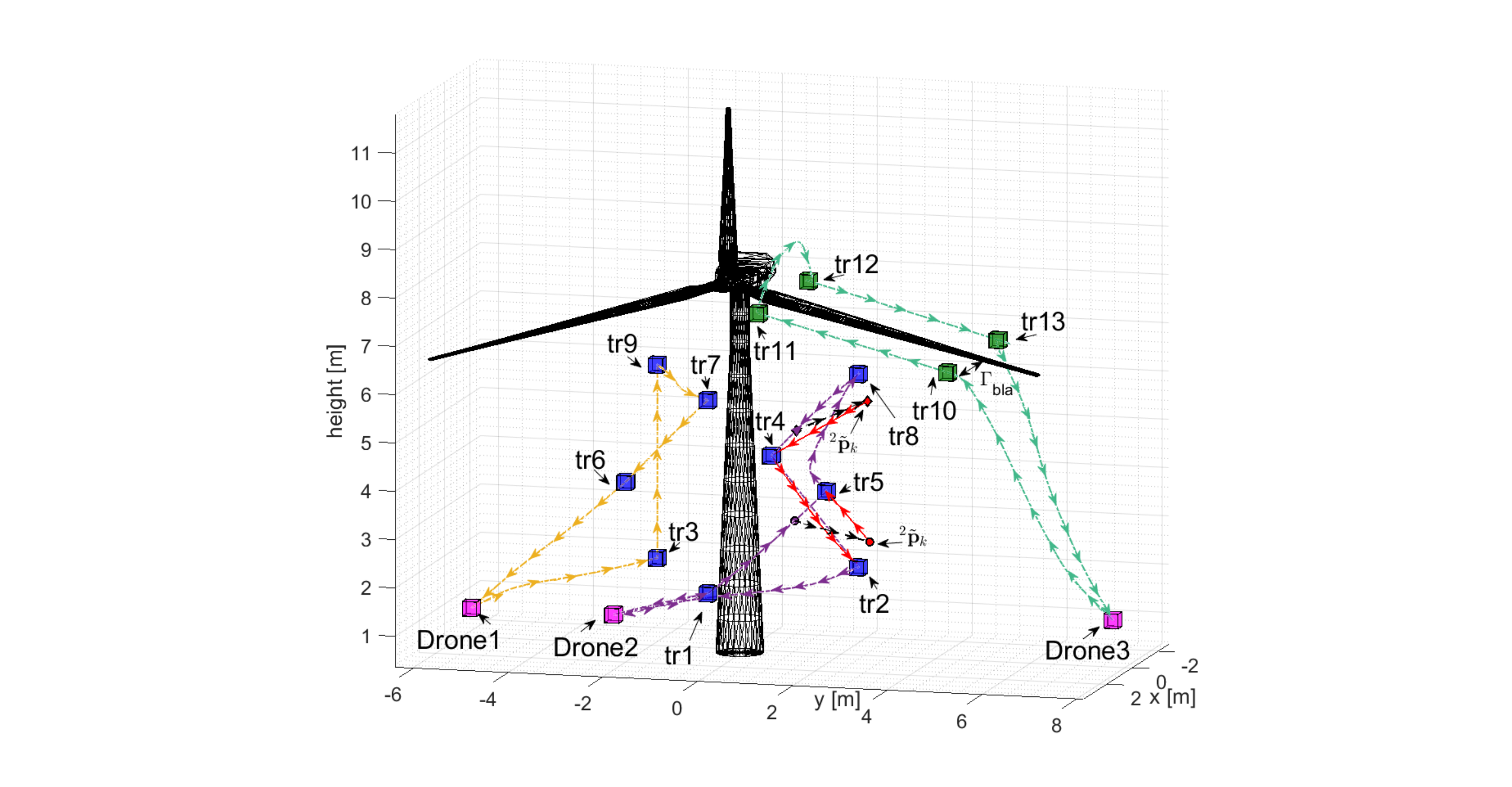}
    \vspace*{-8mm}
    \caption{Event-triggered replanner. Black and red paths represent the deviation from the original path and the updated trajectory from the replanner, respectively.}
    \label{fig:eventTriggeredReplanner}
\end{figure}

This section evaluates the performance of the \textit{attrition-aware} and \textit{event-triggered} replanner. The results of numerical simulations in MATLAB are depicted in Figures~\ref{fig:attritionAwarePlanner} and~\ref{fig:eventTriggeredReplanner}. User preferences were incorporated into the~\ac{STL} robust semantics to demonstrate the importance of safety requirements (${^d}\varphi_\mathrm{dis}$ and ${^d}\varphi_\mathrm{obs}$) in the optimization problem, while also satisfying the same mission criteria. The trajectories obtained with user preferences resulted in safer paths, as shown in Figure~\ref{fig:attritionAwarePlanner}. A generalized robustness score approach was used for the~\ac{STL} robust semantics, which affected the robustness of the safety requirements, as depicted in Figure~\ref{fig:distanceCostFunctionWSTL}. The weights used for the numerical simulations are reported in Table~\ref{tab:tableParamters}. The~\ac{MILP} planning problem~\eqref{eq:MILP} took \SI{4}{\second} to solve, while the~\ac{STL} optimization problem~\eqref{eq:optimizationProblemWeighted} took \SI{107}{\second} to solve. 

To validate the event-triggered replanner's performance, we conducted simulations in the presence of unexpected disturbances that deviated the~\ac{UAV} from its planned path. The replanner was able to detect major deviations (i.e., $\lVert {^d}\tilde{\mathbf{p}}_k - {^d}\mathbf{p}^\star_k \rVert > \eta$) and trigger a partial replanning process online, bringing the~\ac{UAV} back to the next target area, as shown in Figure~\ref{fig:eventTriggeredReplanner}. The planner then checked whether mission requirements were fulfilled and continued replanning until the lost time was recovered. The optimization process took less than \SI{1}{\second} for each replanning.

\subsection{Field experiments}
\label{sec:fieldExperiments}

Experiments were performed with MRS F450 quadrotors~\cite{MRS2022ICUAS_HW, MRS2023JINT_HW} in a mock-up scenario with three~\acp{UAV} and a wind turbine (shown in Figure~\ref{fig:inspectionScenario}). Each~\ac{UAV} was equipped with an Intel NUC computer featuring an i7-8559U processor with $16$GB of RAM, along with the Pixhawk flight controller. The software stack utilized the Noetic Ninjemys release of ROS running on Ubuntu 20.04. For further details, refer to~\cite{MRS2022ICUAS_HW, MRS2023JINT_HW}. The wind turbine was simulated due to safety concerns. The~\acp{UAV} followed trajectories generated in MATLAB and validated in Gazebo, accounting for the presence of the wind turbine. Videos of simulated and experimental results are available at~\url{http://mrs.felk.cvut.cz/milp-stl}. 

\begin{figure}[tb]
    \centering
    \scalebox{0.85}{
    \begin{tikzpicture}
        \node (MultiRobots-Box) at (0,0.2) [fill=gray!3,rounded corners, draw=black!70, densely 
        dotted, minimum height=1.7cm, minimum width=2.75cm]{}; 
        
        \node (MotionPlanner) at (0,0) [text centered, fill=white, draw, rectangle, minimum 
        width=1.5cm, text width=5.5em]{Motion\\Planner};
        
        \draw[-latex] ($(MotionPlanner) - (1.5,0)$) -- node[above]{$\varphi$} (MotionPlanner);
        
        \draw[fill=black] ($ (MotionPlanner.east) + (0.5,0) $) arc(-180:180:0.03);
        \node (GroundStation) at (0,0.75) [text centered]{\small Ground Station};
        
        \node (Drone-Box1) at (5.15,2.15) [fill=gray!3,rounded corners, draw=black!70, densely 
        dotted, minimum height=1.7cm, minimum width=5.45cm]{}; 
        \node (TrackingController1) at (3.65,1.95) [fill=white, draw, rectangle, text centered, 
        text width=5em]{Tracking\\Controller};
        \node (UAVPlant1) at (6.65,1.95) [fill=white, draw, rectangle, text centered, text 
        width=5em]{UAV\\Plant};
        \node (Drone-Box1-Text) at (5.1,2.7) [text centered]{\small $1$\textsuperscript{st} quadrotor};	
        
        \draw[-latex] (TrackingController1) -- node[above]{${^1}\bm{\omega}_c$} node[below]{${^1}T_c$} (UAVPlant1);
        \draw[-latex] (MotionPlanner.east) -- ($ (MotionPlanner.east) + (0.525,0) $) -- ($ (MotionPlanner.east) + (0.525,1.95) $) -- node[above]{\hspace{-0.3em}      ${^1}\mathbf{x}^\star, \hspace{-0.3em} {^1}\mathbf{u}^\star$} node[below]{${^1}\psi$} (TrackingController1.west);
        
        \node (Drone-Box2) at (5.15,0.2) [fill=gray!3,rounded corners, draw=black!70, densely 
        dotted, minimum height=1.7cm, minimum width=5.45cm]{};
        \node (TrackingController2) at (3.65,0) [fill=white, draw, rectangle, text centered, text 
        width=5em]{Tracking\\Controller};
        \node (UAVPlant2) at (6.65,0) [fill=white, draw, rectangle, text centered, text 
        width=5em]{UAV\\Plant};
        \node (Drone-Box2-Text) at (5.15,0.75) [text centered]{\small $2$\textsuperscript{nd} quadrotor};		
        
        \draw[-latex] (TrackingController2) -- node[above]{${^2}\bm{\omega}_c$} node[below]{${^2}T_c$} 
        (UAVPlant2);	 
        \draw[-latex] ($ (MotionPlanner.east) + (0.525,0) $) -- node[above]{\hspace{0.3em}${^2}\mathbf{x}^\star, \hspace{-0.3em} {^2}\mathbf{u}^\star$} node[below]{${^2}\psi$} (TrackingController2.west);
        
        \node (Drone-BoxN) at (5.15,-1.75) [fill=gray!3,rounded corners, draw=black!70, densely dotted, minimum height=1.7cm, minimum width=5.45cm]{};
        \node (Dots2) at (5.15,-0.8) [text centered]{\dots};
        \node (TrackingControllerN) at (3.65,-1.95) [fill=white, draw, rectangle, text centered, text width=5em]{Tracking\\Controller};
        \node (UAVPlantN) at (6.65,-1.95) [fill=white, draw, rectangle, text centered, text width=5em]{UAV\\Plant}; 	
        \node (Drone-BoxN-Text) at (5.15,-1.2) [text centered]{\small $\delta$\textsuperscript{th} quadrotor};
        
        \draw[-latex] (TrackingControllerN) -- node[above]{${^\delta}\bm{\omega}_c$} node[below]{${^\delta}T_c$} (UAVPlantN);
        \draw[-latex] ($ (MotionPlanner.east) + (0.525,0) $) -- ($ (MotionPlanner.east) + (0.525,-1.95) $) -- node[above]{\hspace{0.3em}${^\delta}\mathbf{x}^\star, \hspace{-0.3em}
        {^\delta}\mathbf{u}^\star$} node[below]{${^\delta}\psi$} (TrackingControllerN.west);
        \end{tikzpicture}
    }
    \caption{System Architecture. The ground station's \textit{\ac{STL} Motion Planner} is responsible for generating trajectories (${^1}\mathbf{x}, {^1}\mathbf{u}, \dots, {^\delta}\mathbf{x}, {^\delta}\mathbf{u}$) and heading angles (${^1}\psi, \dots, {^\delta}\psi$) for the multi-rotor \acp{UAV}. These trajectories and heading angles serve as inputs to the \textit{Tracking Controller}, which in turn calculates thrust (${^1}T_c, \dots, {^\delta}T_c$) and angular velocities (${^1}\bm{\omega}_c, \dots, {^\delta}\bm{\omega}_c$) for the~\textit{UAV Plant} \cite{Baca2020mrs}.}
    \label{fig:controlArchitecture}
\end{figure}
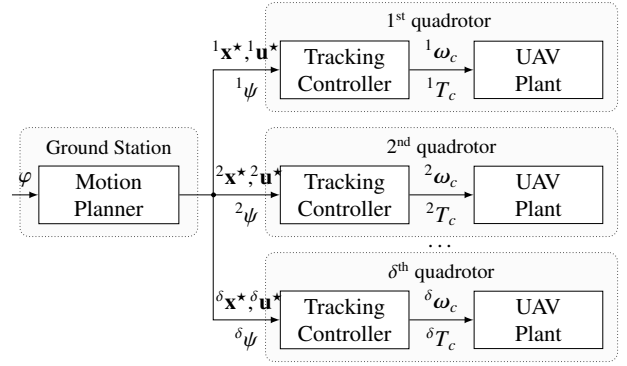

The system architecture, as illustrated in Figure~\ref{fig:controlArchitecture}, integrates the~\ac{STL} motion planner responsible for solving the optimization problem~\eqref{eq:optimizationProblemMotionPrimitives} to generate trajectories (${^d}\mathbf{x}^\star, {^d}\mathbf{u}^\star$) and heading angles (${^d}\psi$) for the fleet. This trajectory generation process occurs as a one-shot computation at time $t_0$, with the resulting trajectories serving as references for the~\ac{UAV} trajectory tracking controller~\cite{Baca2020mrs}.

During the flight tests, we confirmed the successful completion of the inspection mission specified by the~\ac{STL} formula~\eqref{eq:windTurbineInspectionFormula}. These flights also demonstrated adherence to physical constraints and safety features, including velocity (${^d}\underline{v}$ and ${^d}\bar{v}$) and acceleration (${^d}\underline{a}$ and ${^d}\bar{a}$) constraints, minimum safety distance ($\Gamma_\mathrm{dis}$), and blade distance ($\Gamma_\mathrm{bla}$) detailed in Table~\ref{tab:tableParamters}. Figure~\ref{fig:experimentsTimeLine} provides snapshots from the experiments, illustrating the~\acp{UAV}' proximity to wind turbine infrastructure such as pylons and blades. To visualize this, we utilized a ROS package to project 3D mesh files -- originally utilized in MATLAB and Gazebo simulations -- onto the camera frames of the~\acp{UAV}.

\subsection{Discussion}
\label{sec:discussionPreviousResult}

In this discussion section, we aim to compare our earlier work in~\cite{SilanoRAL2021} with the current study, highlighting significant improvements and advantages.

Firstly, the current solution yields feasible trajectories with diverse time bounds and vehicle constraints, while the approach proposed in~\cite{SilanoRAL2021} assumes quadrotors have the same physical constraints in terms of maximum velocity and acceleration ($\{ \lvert {^d}\underline{v}^{(j)} \rvert = \lvert {^d}\bar{v}^{(j)} \rvert, \lvert {^d}\underline{a}^{(j)} \rvert = \lvert {^d}\bar{a}^{(j)} \rvert, \forall d \in \mathcal{D} \}$). Furthermore, and more importantly, the inspection mission in~\cite{SilanoRAL2021} is limited to visiting target areas of interest (for taking photos of power line insulators and tower mechanical structures), similar to the pylon inspection specification. However, in the pylon inspection, target areas are assigned to the drones by the~\ac{MILP} \eqref{eq:MILP}, so one drone could theoretically cover all targets while the second could remain on the ground. In the algorithm proposed in~\cite{SilanoRAL2021}, the targets are manually assigned to the drones. Finally, the optimization problem in \eqref{eq:optimizationProblemMotionPrimitives} sets a minimum robustness threshold value ($\tilde{\rho}_\varphi({^d}\mathbf{p}^{(j)}, {^d}\mathbf{v}^{(j)}) \geq \zeta$) that acts as a safety buffer, ensuring the satisfaction of the~\ac{STL} formula $\varphi$ even in the presence of disturbances, which is not included in the previous work.

Secondly, the collaborative mission requirements, along with the heterogeneous constraints, make the problem challenging to solve. The closer the initial guess is to the optimal solution, the higher the chances that the nonlinear optimization problem will be feasible and converge to the optimal solution. This is the second contribution of the paper, i.e., proposing a two-step hierarchical solution for computing the initial solution.

Lastly, as discussed in the comparative analysis between the~\ac{MILP} and the~\ac{STL} solution seeded with the~\ac{MILP} initial guess in Section~\ref{sec:comparativeAnalysis}, the pure~\ac{STL} approach (as structured in the authors' previous work~\cite{SilanoRAL2021}) does not converge with a generic initial condition. This underscores the importance of having an~\ac{MILP} problem to warm-start the~\ac{STL} solution computation. This is another significant contribution of the manuscript.



\section{Conclusions}
\label{sec:conclusions}

This paper has presented a motion planning framework for collaborative inspection missions using a fleet of multi-rotor~\acp{UAV} under heterogeneous constraints, with a focus on wind turbine inspection. The approach uses~\ac{STL} specifications to generate feasible trajectories meeting mission requirements, including safety and mission time requirements. An~\ac{MILP} approach provides a feasible initial guess solution for the~\ac{STL} planner, which helps solution convergence. An event-triggered replanning and attrition-aware planning handle failure and conflicting tasks. Validation has been performed through MATLAB and Gazebo simulations and field experiments. 

Our investigation has revealed that depending solely on the simplified \ac{MILP} solver falls short for our application. Nonetheless, it does serve as a valuable foundation for the complete \ac{STL} planner. Embracing a hierarchical approach empowers us to handle a broader range of mission specifications and requirements compared to existing methods, albeit with a rise in computational complexity. In future work, we plan to investigate risk-aware techniques to model sensor failure and communication dropouts in the planning problem. Additionally, delving into conflicting temporal logic specifications and other temporal logic languages will expand the framework's utility to dynamically changing environments.



\section*{Acknowledgments}
\label{sec:Acknowledgments}

The authors would like to thank Daniel Smrcka, Jan Bednar, Jiri Horyna, Tomas Baca, and the MRS group in Prague for their help with the field experiments. This publication is part of the R+D+i project TED2021-131716B-C22, funded by MCIN/AEI/10.13039/501100011033 and by the European Union NextGenerationEU/PRTR. This work was also partially supported by the European Union's Horizon 2020 research and innovation project AERIAL-CORE under grant agreement no. 871479, by the ECSEL Joint Undertaking (JU) research and innovation programme COMP4DRONES under grant agreement no. 826610, by the Czech Science Foundation (GAČR) grant no. 23-07517S, by CTU grant no. SGS23/177/OHK3/3T/13, and by the European Union under the project Robotics and Advanced Industrial Production (reg. no. CZ.02.01.01/00/22 008/0004590).



\bibliographystyle{elsarticle-num}
\bibliography{bib}



\vspace{1em}

\begin{wrapfigure}{l}{25mm} 
    \includegraphics[width=1in,height=1.5in,clip,keepaspectratio]{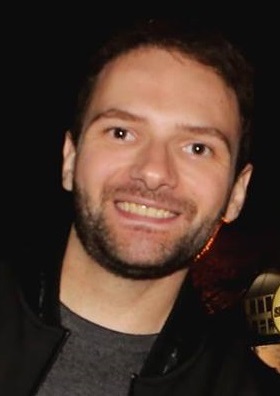}
\end{wrapfigure}\par
\textbf{Giuseppe Silano} (Member, IEEE) is a tenured researcher at Ricerca sul Sistema Energetico and an associated researcher at Czech Technical University in Prague (CTU-P). He earned his B.Sc., M.Sc., and Ph.D. from the University of Sannio, Italy, in 2012, 2016, and 2020, respectively. During his Ph.D., he visited LAAS-CNRS and participated in the MBZIRC 2020 robotic competition. He was a post-doctoral researcher at CTU-P with the MRS group (2020-2022) and a visiting researcher at the RAM group, University of Twente, in 2022. His research focuses on UAV motion planning, model predictive control, formal methods for robotics, communication-aware robotics, and human-robot collaboration. He has authored over 30 publications and held leadership roles in European research projects. Since 2022, he serves as an associate editor for key conferences. \par

\vspace{1em}

\begin{wrapfigure}{l}{25mm} 
    \includegraphics[width=1in,height=1.5in,clip,keepaspectratio]{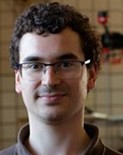}
\end{wrapfigure}\par
\textbf{Alvaro Caballero} earned his B.Sc. in aerospace engineering, M.Sc. in aeronautical engineering, and Ph.D. in aerial robotics from the University of Seville, Spain, in 2014, 2016, and 2022, respectively. Currently, he holds a post-doctoral position at the GRVC Robotics Lab, University of Seville. Since 2014, he has contributed to various projects, including FP7 EC-SAFEMOBIL, MBZIRC 2017, and H2020 AEROARMS, HYFLIERS, AERIAL-CORE, and OMICRON. He also collaborates with industry leaders like NAVANTIA or ENEL. His main research interests focus on motion planning for aerial manipulation in inspection and maintenance operations.\par

\vspace{1em}

\begin{wrapfigure}{l}{25mm} 
    \includegraphics[width=1in,height=1.5in,clip,keepaspectratio]{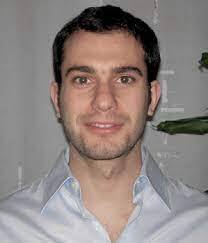}
\end{wrapfigure}\par
\textbf{Davide Liuzza} obtained his Ph.D. in automation engineering from the University of Naples Federico II, Italy, in 2013. He conducted research visits to institutions in the UK and Sweden during his doctoral studies. Afterward, he held postdoctoral positions at KTH and the University of Sannio. Subsequently, he was a visiting researcher at Chalmers University of Technology and a staff researcher at ENEA. Currently, he serves as an Assistant Professor at the University of Sannio. His research focuses on networked control systems, multiagent system coordination, nonlinear systems' stability, and energy system control. He also explores topics like human-robot coordination and real-time control of nuclear fusion systems.\par

\vspace{1em}

\begin{wrapfigure}{l}{25mm} 
    \includegraphics[width=1in,height=1.5in,clip,keepaspectratio]{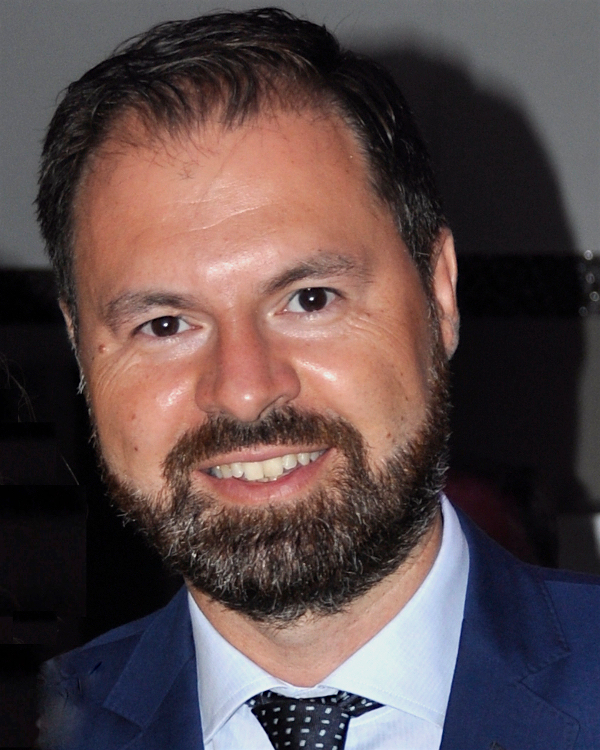}
\end{wrapfigure}\par
\textbf{Luigi Iannelli} (Senior Member, IEEE) earned his M.Sc. in computer engineering from the University of Sannio, Italy, in 1999, and his Ph.D. in information engineering from the University of Napoli Federico II, Italy, in 2003. He became an associate professor of automatic control at the University of Sannio in 2016 after serving as an assistant professor. With research visits to institutions in Sweden and the Netherlands, his work centers on analyzing and controlling switched systems, stability of piecewise-linear systems, smart grid control, and applying control theory to power electronics and UAVs. He co-edited ``Dynamics and Control of Switched Electronic Systems'' (Springer, 2012).\par

\vspace{1em}

\begin{wrapfigure}{l}{25mm} 
    \includegraphics[width=1in,height=1.5in,clip,keepaspectratio]{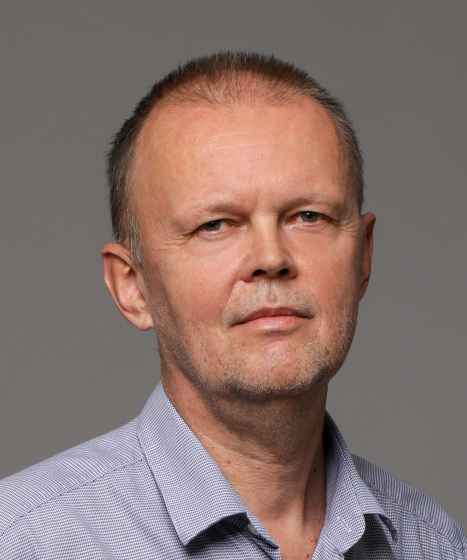}
\end{wrapfigure}\par
\textbf{Stjepan Bogdan} (Senior Member, IEEE) earned his B.Sc., M.Sc., and Ph.D from the University of Zagreb, Croatia, in 1990, 1993, and 1999, respectively. He was a Fulbright Researcher at the Automation and Robotics Research Institute, Arlington, USA, under Prof. Frank Lewis. Currently, he is a Full Professor at the Laboratory for Robotics and Intelligent Control Systems (LARICS), University of Zagreb. Bogdan has coauthored four books and numerous articles on topics such as autonomous systems, aerial robotics, multi-agent systems, intelligent control systems, bio-inspired systems, and discrete event systems.\par

\vspace{1em}

\raggedend

\begin{wrapfigure}{l}{25mm} 
    \includegraphics[width=1in,height=1.5in,clip,keepaspectratio]{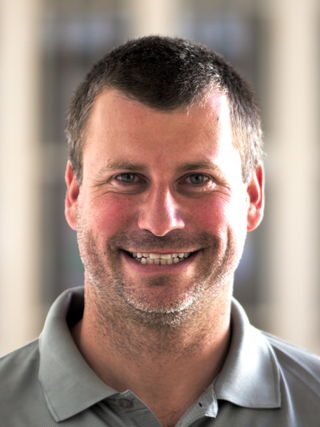}
\end{wrapfigure}\par
\textbf{Martin Saska} (Member, IEEE) holds an M.Sc. from Czech Technical University (2005) and a Ph.D. from the University of Wuerzburg, Germany. He was a Visiting Scholar at the University of Illinois at Urbana-Champaign (2008) and the University of Pennsylvania (2012, 2014, and 2016). Since 2009, he has been with the Czech Technical University, first as a research fellow and then as an associate professor, leading the Multi-Robot Systems Lab and co-founding the Center for Robotics and Autonomous Systems. Saska has authored/coauthored more than 90 peer-reviewed conference papers and 60 journal publications.\par


\end{document}